\documentclass[electronic]{vgtc}             

\ifpdf
  \pdfoutput=1\relax                   
  \usepackage{graphicx}                
  \DeclareGraphicsExtensions{.pdf,.png,.jpg,.jpeg} 
\else
  \usepackage{graphicx}                
  \DeclareGraphicsExtensions{.eps}     
\fi%

\graphicspath{{figures/}{pictures/}{images/}{./}} 

\usepackage{microtype}                 
\PassOptionsToPackage{warn}{textcomp}  
\usepackage{textcomp}                  
\usepackage{amsmath}
\usepackage{mathptmx}                  
\usepackage{times}                     
\usepackage{cite}                      
\usepackage{tabu}                      
\usepackage{booktabs}                  
\usepackage{subfigure}
\usepackage{algorithm}
\usepackage{algpseudocode}
\usepackage{hyperref}
\usepackage[T1]{fontenc}

\onlineid{0}

\vgtccategory{Research}

\vgtcinsertpkg

\newcommand{\sysname}{VIO-APR\space}

\title{Robust Localization with Visual-Inertial Odometry Constraints for Markerless Mobile AR}

\author{Changkun Liu\thanks{e-mail: cliudg@connect.ust.hk}\\ %
        \scriptsize HKUST %
\and Yukun Zhao\thanks{e-mail: yzhaoeg@connect.ust.hk}\\ %
     \scriptsize HKUST %
\and Tristan Braud\thanks{e-mail: braudt@ust.hk}\\ %
     \scriptsize HKUST}

\abstract{%
Visual Inertial Odometry (VIO) is an essential component of modern Augmented Reality (AR) applications. However, VIO only tracks the relative pose of the device, leading to drift. Absolute pose estimation methods infer the device's absolute pose, but their accuracy depends on the input quality.
This paper introduces VIO-APR, a new framework for markerless mobile AR that combines an absolute pose regressor (APR) with a local VIO tracking system. VIO-APR uses VIO to assess the reliability of the APR and the APR to identify and compensate for VIO drift. This feedback loop results in more accurate positioning and more stable AR experiences.
To evaluate VIO-APR, we created a dataset that combines camera images with ARKit's VIO system output for six indoor and outdoor scenes of various scales.
Over this dataset, VIO-APR improves the median accuracy of popular APR by up to 36\% in position and 29\% in orientation, increases the percentage of frames in the high ($0.25 m, 2^{\circ}$) accuracy level by up to 112\% and greatly reduces the percentage of frames predicted below the low ($5 m, 10^\circ$) accuracy level. We implement VIO-APR into a mobile AR application using Unity to demonstrate its capabilities. VIO-APR results in fast inference (<100ms)  with a noticeably more accurate and stable overall experience.

} 

\CCScatlist{
  \CCScatTwelve{Visual Localization}{Camera pose regression}{Visual-Inertial Odometry}{Visual positioning system};
  \CCScatTwelve{Computer Vision}{Augmented Reality}{}{}
}

\begin{document}



\maketitle
\section{Introduction}

Augmented reality (AR) applications increasingly rely on visual-inertial odometry (VIO) to track the device's position and orientation. 
However, VIO systems can only track the device's pose in a local coordinate system and do not provide the absolute device pose in a shared world coordinate system. As such, most AR applications still rely on 2D markers to anchor digital content in the physical world.

Absolute camera pose estimation would allow anchoring digital content in the physical world without markers, 
paving the way to ubiquitous AR. Although the computer vision community has made significant progress in this field, many challenges hinder its application to AR. Most absolute pose estimation techniques rely  on structure-based methods that detect and match visual features in the images against a 3D model of the environment or absolute pose regressors (APR) that  regress the camera pose with a single monocular image input through end-to-end machine learning models. 
Structure-based localization methods have achieved great accuracy in pose estimation~\cite{dusmanu2019d2,sarlin2019coarse,taira2018inloc,noh2017large}. 
However, they are often computationally intensive, taking hundreds of milliseconds to provide results on a GPU-equipped server.  They also require storing a large environment model. As such, structure-based localization is only considered in offloaded AR scenarios where end-to-end latency is not critical.
APR methods make decisions at least one order of magnitude faster than structure-based methods. 
However, they offer lower median accuracy and are more prone to large errors (over than 5 meters and 10 degrees), especially in large outdoor environments. Such imprecision can lead to severe alignment problems and unreliable AR experiences.
Due to the significant issues with both strategies, 2.5D localization has become popular in the AR community~\cite{arth2015instant}. This technique matches the general shape of the environment (typically buildings) against a 2.5D map with limited features. However, although this technique can yield fast and accurate results in urban or indoor environments that are easy to represent as 2.5D maps, they often require additional data, such as building outlines or dimensions. Additionally, 2.5D localization primarily focuses on urban environments and buildings, limiting its applicability.


In this paper, we argue that the fast inference speed and low storage overhead of absolute pose regressor (APR) methods outweigh their constraints for real-time visual localization on mobile devices like AR. AR applications involve significant mobility and require periodic absolute pose estimation to recalibrate the local tracking mechanisms such as inertial measurement units or simultaneous localization and mapping (SLAM)~\cite{yu2022improving, bao2022robust}. Achieving high accuracy on a few selected frames is more critical than localizing all frames in this context.
We design a new framework that improves APR's robustness with easily accessible VIO data from mobile devices.
APR output tends to present a high variability around the ground truth pose. Modern VIO systems track the pose very accurately over small displacements at the cost of increasing drift. \sysname, our proposed framework, leverages these properties to reduce the noisiness of the APR pose while minimizing the impact of VIO drift across the experience. The VIO system also tracks the device's pose between absolute pose estimations, reducing the need for frequent relocalizations.
To evaluate our pipeline, we release a new dataset that includes both outdoor and indoor scenes with raw sensor data accessible on a modern-day smartphone. The dataset includes the VIO pose of each frame as provided by ARKit.
We implement \sysname over PoseNet (PN)~\cite{kendall2015posenet,kendall2017geometric} and MS-Transformer (MS-T)~\cite{shavit2021learning}, two popular APR methods. Over this dataset, \sysname significantly improves the accuracy compared to solely using the corresponding APR. We also implement \sysname into an AR application that directly leverages ARKit data to optimize the APR poses. \sysname results in more accurate localizations with a lower variability, and thus smoother situated AR experiences. 

We summarize our main contributions as follows:
\begin{enumerate}
\item   We \textbf{design an APR-agnostic framework}, \sysname, fusing information from a mobile VIO system to model the uncertainty of APR outputs. We select poses with low uncertainty and filter poses with high uncertainty. Reference poses are calculated from the selected poses to optimize some estimated poses from APR.

\item To evaluate our approach, we release a \textbf{rich indoor and outdoor dataset with VIO data}. Each image in the dataset has multiple pose labels, including ground truth from COLMAP and the 6DoF pose label from ARKit. We also analyze the absolute and relative pose error between ARkit and SfM poses.

\item We \textbf{implement and evaluate \sysname} over two APR models. With \sysname, both models significantly improve accuracy and all precision levels. Our framework improves the accuracy of MS-T as much as 36\% on position error and 29\% on orientation error in median error over the average of all three outdoor scenes.
\item We integrate \sysname into a \textbf{mobile AR application}. \sysname results in more accurate relocalizations and smoother AR experiences with a low computation footprint (100\,ms on-device localization time) and low energy consumption. 
\end{enumerate}


\section{Related work}
This paper aims to improve the accuracy of APR methods by using local VIO for keyframe selection and pose optimization. In this section, we first review the most relevant works in the field of APR before discussing methods for keyframe selection and pose optimization. Finally, we review the most common datasets used for evaluating visual positioning algorithms.

\subsection{Absolute Pose Regression}
Absolute Pose Regressors train Deep Neural Networks for regression to directly predict the 6-DOF camera pose including position and orientation of a query image. The seminal work in this area is introduced by Kendall et al. in PoseNet~\cite{kendall2015posenet}. Since then, there have been several improvements to APR, mainly in terms of modifying the architecture and loss function of the neural network~\cite{kendall2016modelling,kendall2017geometric,melekhov2017image,wu2017delving,shavit2021learning,chen2021direct,chen2022dfnet} and using auxiliary learning~\cite{radwan2018vlocnet++,valada2018deep,brahmbhatt2018geometry,moreau2022coordinet}. 
Some works use unlabeled data to finetune APR~\cite{brahmbhatt2018geometry, chen2022dfnet} while others increase the training data size with synthetic images~\cite{naseer2017deep,wu2017delving,sattler2019understanding,moreau2022lens}.

MapNet~\cite{brahmbhatt2018geometry} is the closest work to our study, as it minimizes the loss of the per-image absolute pose and the loss of the relative pose between image pairs. MapNet+ is a version of MapNet augmented with unlabeled data in a semi-supervised manner. However, MapNet relies on GPS interpolation data and DSO to compute the relative pose for each image pair~\cite{engel2017direct} as no VIO dataset was available then. Our work extends such an idea to an APR-agnostic pipeline using commercially available VIO systems such as ARKit to improve the accuracy of the absolute pose estimation. 
To evaluate the performance of our pipeline, we collect a dataset with high-end mobile devices that integrates sensor and VIO data for each camera frame, together with SfM ground truth (GT) poses.  Our pipeline adds negligible extra overhead compared to time-consuming fine-tuning methods and does not require additional unlabeled test data that may be difficult to acquire in real-life applications~\cite{chen2023refinement}.

\subsection{Keyframe Selection and Pose Optimization}

Keyframe selection is typical in Simultaneous Localization and Mapping (SLAM)~\cite{tan2013robust,mur2015orb,mur2017orb}, where unique feature points are identified for relocalization and loop closing. 
APR methods are tend to overfit their training data~\cite{sattler2019understanding}. Keyframe selection for APR aims to infer which images will likely result in accurate pose estimation.
Uncertainty estimation is able to find the outliers.
Bayesian PoseNet\cite{kendall2016modelling}, AD-PoseNet~\cite{huang2019prior}, BMDN\cite{deng2022deep,bui20206d}, VaPoR\cite{zangeneh2023probabilistic} and CoordiNet~\cite{moreau2022coordinet} can output poses and uncertainty simultaneously, but they are less accurate than many APR methods that only provide estimated poses. Besides, they rely on a specific neural network structure and loss functions to model uncertainty by estimating the variance of the pose distribution. In contrast,  our pipeline directly compares APR's prediction's relative pose distance with the mobile VIO's relative pose distance at test time. This approach is more interpretable in real-life scenarios~\cite{sattler2019understanding}, enabling its integration into real-time AR applications.


Several methods aim to optimize the absolute pose. MapNet+PGO~\cite{brahmbhatt2018geometry} uses a constant Pose Graph Optimization (PGO) method. However, it tends to optimize well-predicted poses negatively. AD-PoseNet~\cite{huang2019prior} proposes an uncertainty-aware PGO that only optimizes unreliable poses using multiple images. 

Zhang \emph{et al.}\cite{zhang2021reference}, Taira \emph{et al.}\cite{taira2018inloc}, and Chen \emph{et al.}~\cite{chen2023refinement} verify whether the estimated pose is reliable by rendering synthetic image based on the estimated pose.  Iterative algorithms are used to minimize the gap between the query and synthetic images at test time.  These methods discards some of the advantages of APR such as the low computation and storage footprint, making APR less efficient than typical structure-based methods. Our method performs a rigid transformation of the VIO's pose to optimize unreliable APR poses directly, without iterative optimization. It establishes a feedback loop that also compensates for VIO drift. 
Additionally, our framework is flexible and APR-agnostic, allowing both uncertainty-aware and uncertainty-unaware pre-trained APR models to be used without modifying training details. This approach improves the accuracy of traditional methods with minimal overhead, enabling reliable APR usage in mobile AR applications.




\subsection{Camera localization datasets}

Benchmark datasets of visual localization provide camera poses for a set of training and test images. They are commonly used to evaluate performance in autonomous robots~\cite{lim2012real}, self-driving cars~\cite{heng2019project}, and Augmented Reality (AR) systems~\cite{arth2011real,castle2008video,lynen2015get}. 


Cambridge dataset~\cite{kendall2015posenet} and 7Scenes~\cite{glocker2013real,shotton2013scene} are the most popular image-based localization dataset with SfM GT. The GT of 7Scenes are originally obtained from RGB-D SLAM (dSLAM), and~\cite{chen2023refinement} extends it with SfM tool COLMAP~\cite{schoenberger2016sfm} for higher accuracy proved by ~\cite{brachmann2021limits}. These two datasets present typical indoor and outdoor locations that translate well to the AR context. However, they only offer visual data. Many other vision-only datasets have been developed. However, such datasets present significant limitations in terms of scope ~\cite{sattler2018benchmarking, valentin2016learning, sturm2012benchmark,glocker2013real,shotton2013scene}, hardware~\cite{glocker2013real,shotton2013scene,valentin2016learning,sturm2012benchmark,taira2018inloc}, and environment size~\cite{sturm2012benchmark, glocker2013real,shotton2013scene,taira2018inloc}. Datasets that provide additional sensor data alongside visual data often lack the crucial VIO data required to assess our pipeline~\cite{kendall2015posenet, do2022learning, glocker2013real,shotton2013scene,taira2018inloc, maddern20171,sattler2018benchmarking}.

To the best of our knowledge, only two datasets consider  VIO data from mobile devices such as ARKit or ARCore. LaMAR\cite{sarlin2022lamar} is a dataset specifically developed for AR localization tasks. However, despite optimizing pseudo-GT global poses by fusing LiDAR, SfM, wireless, and VIO data, Lamar only provides the pseudo-GT, and other sensor data is not public.  ADVIO\cite{cortes2018advio} obtains GT by combining inertial navigation with manual position fixes using iPhone 6S and Tango. However, both platforms are nowadays obsolete. Tango was discontinued for ARCore and each new version of ARKit significantly reduces drift~\cite{scargill2022here}. No dataset reflects the current state of VIO tracking for the AR community. This paper thus introduces a new dataset captured using iPhone 14 Pro Max, the flagship ARKit 6 phone at the time. We additionally provide SfM GT and the pose labels of VIO, IMU, GPS, and compass in their respective coordinates. We also use this dataset to characterize the position and orientation drift of ARKit over multiple scenes.
Our dataset highlights the low drift of current mobile VIO solutions, such as ARKit, supporting our assumption that VIO can reinforce absolute pose estimation. It enables researchers to consider multiple sensor and VIO data for localization tasks in AR.

\section{Proposed Approach}
\label{sec:method}

\subsection{Definition}
 Given a stream of query images $I_{i}$,  APR $\mathcal{R}$ outputs estimated global position $\mathbf{\hat{x}}_i$ and rotation $\mathbf{\hat{q}}_i$ so that $\mathcal{R}({I}_i) = \hat{p}_i =  <\mathbf{\hat{x}_i},\mathbf{\hat{q}}_i>$ is the 6Dof camera pose for each image. The camera pose of each image $I_{i}$ in the VIO coordinate system is noted $p_i^{vio} = <\mathbf{x}_i^{vio},\mathbf{q}_i^{vio}>$. $p_i = <\mathbf{x}_i,\mathbf{q}_i>$ is the GT pose of each image $I_{i}$.  The relative translation between image $I_{i}$ and $I_{i+1}$ is characterized by
 \begin{equation}
     \hat{\Delta}_{trans} (i+1,i) = ||\mathbf{\hat{x}}_{i+1}-\mathbf{\hat{x}}_i||_2,
 \end{equation}
 \begin{equation}
     \Delta_{trans}^{vio}(i+1,i) = ||\mathbf{x}^{vio}_{i+1}-\mathbf{x}^{vio}_i||_2
 \end{equation}
 \begin{equation}
     \Delta_{trans}(i+1,i) = ||\mathbf{x}_{i+1}-\mathbf{x}_i||_2
 \end{equation}
Similarly, we get relative rotation in degree follow~\cite{hartley2013rotation}, $\mathbf{q}^{-1}$ denotes the conjugate of $\mathbf{q}$, and we assume all quaternions are normalized:
\begin{equation}
     \hat{\Delta}_{rot}(i+1,i) = 2\arccos{|\mathbf{\hat{q}}_{i+1}^{-1}\mathbf{\hat{q}}_{i}|}\frac{180}{\pi}
\end{equation}
\begin{equation}
     \Delta_{rot}^{vio}(i+1,i) = 2\arccos{|\mathbf{q}_{i+1}^{vio-1}\mathbf{q}_{i}^{vio}|}\frac{180}{\pi}
\end{equation}
 \begin{equation}
     \Delta_{rot}(i+1,i) = 2\arccos{|\mathbf{q}_{i+1}^{-1}\mathbf{q}_{i}|}\frac{180}{\pi}
 \end{equation}
 $\hat{u}_{i,i+1}$ is the odometry of $I_{i}$ and $I_{i+1}$ from predicted poses of APR. $u_{i,i+1}^{vio}$ is the odometry of $I_{i}$ and $I_{i+1}$ from the VIO system. $u_{i,i+1}$ is the GT odometry of $I_{i}$ and $I_{i+1}$. The definition of $\hat{u}_{i,i+1}$, 
$u_{i,i+1}^{vio}$ and $u_{i,i+1}$ is shown as below,
 \begin{equation}
     \hat{u}_{i,i+1} = <\hat{\Delta}_{trans}(i,i+1), \hat{\Delta}_{rot}(i,i+1)>
\end{equation}

\begin{equation}
     u_{i,i+1}^{vio} = <\Delta_{trans}^{vio}(i,i+1), \Delta_{rot}^{vio}(i,i+1)> 
 \end{equation}

 \begin{equation}
     u_{i,i+1} = <\Delta_{trans}(i,i+1), \Delta_{rot}(i,i+1)> 
 \end{equation}

We define the Relative Position Error (RPE) and the Relative Orientation Error (ROE) for the VIO and the APR as follows:

\begin{equation}
    \text{RPE}_{<vio,GT>}  = |\Delta_{trans}(i+1,i) - \Delta_{trans}^{vio}(i+1,i)|
\label{eq:rpe}
\end{equation}

\begin{equation}
     \text{ROE}_{<vio,GT>} = |\Delta_{rot}(i+1,i) - \Delta_{rot}^{vio}(i+1,i)|
\label{eq:roe}
\end{equation}

\begin{equation}
    \text{RPE}_{<apr,vio>}  = |\hat{\Delta}_{trans}(i+1,i) - \Delta_{trans}^{vio}(i+1,i)|
\end{equation}

\begin{equation}
     \text{ROE}_{<apr,vio>} = |\hat{\Delta}_{rot}(i+1,i) - \Delta_{rot}^{vio}(i+1,i)|
\label{eq:roe_apr_vio}
\end{equation}

\subsection{Detecting accurate pose estimations using VIO}
\label{sec:detect}
As shown in Section~\ref{sec:ds}, modern VIO systems have low drift at a small temporal scale, and $u_{i,i+1}^{vio}$ tends to be very close to the GT as shown in Figure~\ref{fig:errdis}. More than $90\%$ of the images pairs have less than $0.1m$  of $\text{RPE}_{<vio,GT>}$ and less than $1^\circ$ of $\text{ROE}_{<vio,GT>}$. 
Therefore, we can assume that $u_{i,i+1}^{vio}$ is almost the GT odometry for frame $I_{i+1}$, $u_{i,i+1}$. 
We model the uncertainty of the APR output, $\hat{p}_{i+1}$, with $u_{i,i+1}^{vio}$, taking advantage of the relationship between $\hat{u}_{i,i+1}$  and $u_{i,i+1}^{vio}$. If the $\text{RPE}_{<apr,vio>}$ and $\text{ROE}_{<apr,vio>}$ of multiple consecutive images are very small, we consider the predicted poses to be accurate. 

We define a distance threshold $d_{th}$ for $\text{RPE}_{<apr,vio>}$  and an orientation threshold $o_{th}$ for $\text{ROE}_{<apr,vio>}$.
An estimated pose from the APR is considered accurate if it is close to GT with the error less than  $\frac{d_{th}}{2}$ and $\frac{o_{th}}{2}$. Given two consecutive query images $I_{i}$, $I_{i+1}$,
\begin{enumerate}
    \item Estimated poses of $I_{i}$ and $I_{i+1}$ are accurate, then $\text{RPE}_{<apr,vio>}$  and $\text{ROE}_{<apr,vio>}$ are lower than $d_{th}$ and $o_{th}$, respectively.
    \item One of the estimated pose for $I_{i}$ and $I_{i+1}$ is not accurate, then either $\text{RPE}_{<apr,vio>}$ should be larger than $d_{th}$ or $\text{ROE}_{<apr,vio>}$ should be larger than $o_{th}$.
    \item  Both estimated poses of $I_{i}$ and $I_{i+1}$ are inaccurate. However, $\text{RPE}_{<apr,vio>}$  remains lower than $d_{th}$ and $\text{ROE}_{<apr,vio>}$ lower than $o_{th}$.
    \item Both estimated poses of $I_{i}$ and $I_{i+1}$ are inaccurate, and either $\text{RPE}_{<apr,vio>}$ is larger than $d_{th}$ or $\text{ROE}_{<apr,vio>}$ is larger than $o_{th}$.
\end{enumerate}

When either $\text{RPE}_{<apr,vio>}$ or $\text{ROE}_{<apr,vio>}$ is larger than its respective threshold (case (2) and (4)), the pose is flagged as inaccurate and can thus be filtered out. Similarly, in case (1), the two poses are identified as accurate.
In case (3), two inaccurate poses are identified as accurate. Our method uses a probabilistic approach to reducing such false positives. APR error tends to be random with a large variance. As such, two consecutive images presenting a large APR error while being close to each other in the same direction as the VIO is a rare occurrence. By comparing more pairs of images, we further reduce the probability of false positive, filtering out the most unreliable predictions.
We then obtain the rigid transformation between the VIO coordinate system and the global coordinate system by using the reliable predicted poses and VIO poses. To ensure the rigid transform relationship's reliability, we calculate the average pose of selected predicted poses as reference pose.  The rotation and translation of the coordinate system of VIO and the world coordinate system change over time due to the VIO drift. Therefore, we only need to update the reliable poses occasionally and optimize the predicted pose by calculating the new rotation and translation.

\begin{figure*}[!t]
  \centering
  \includegraphics[width=0.8\linewidth]{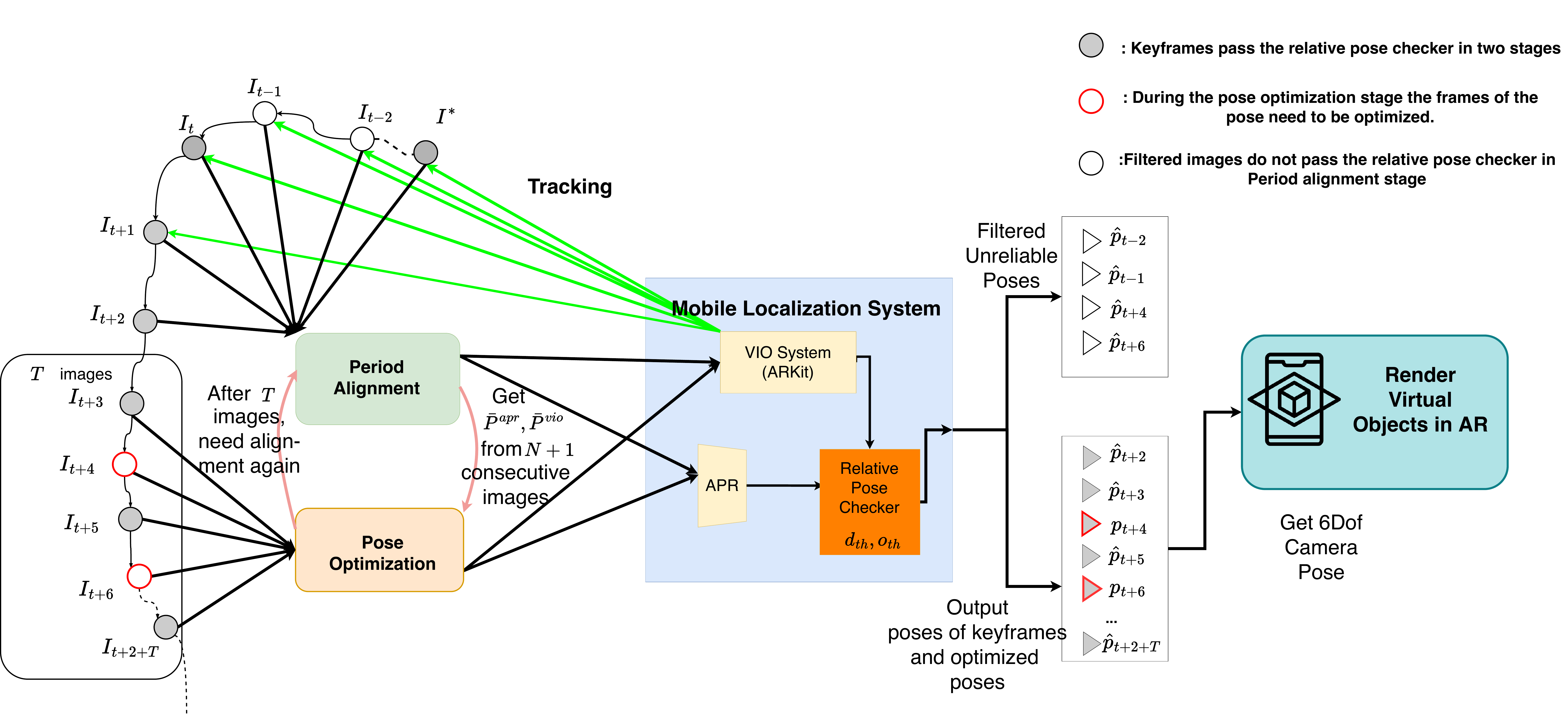}
\caption{ \sysname framework in a mobile AR system. We set $N = 2$. $I^*$ is the keyframe from the last pose optimization stage. We can only confirm $I_{t}$ and $I_{t+1}$ are keyframes after obtaining the predicted pose of $I_{t+2}$. 
Therefore, we use local tracking between the last reliable pose from $I^*$ until finding the new reference pose $I_{t+2}$.
 We calculate the reference pose $\Bar{P}^{apr}$ based on $\hat{p}_{t},\hat{p}_{t+1},\hat{p}_{t+2}$. We calculate the reference pose $\Bar{P}^{vio}$ based on $p_{t}^{vio},p_{t+1}^{vio},p_{t+2}^{vio}$. 
 We calculate the rigid transformation between $\Bar{P}^{apr}$ and $\Bar{P}^{vio}$ to optimize unreliable poses in optimazition stage.
 }
\label{fig:framework}
\end{figure*}

\subsection{\sysname framework} 
\label{subsec:vio-apr}
Based on the above subsections,  we present \sysname a new APR-agnostic framework that combines the outputs of APR methods and information from smartphones' VIO systems to improve pose prediction accuracy. 
The framework keeps the most reliable prediction poses with the help of VIO, identifies unreliable poses, and optimizes them based on reliable poses. 
Our framework combines the distinctive features of APR predictions, which are locally noisy but drift-free, with mobile VIO systems, which are locally smooth but tend to drift, as shown by~\cite{brahmbhatt2018geometry}. 
This framework consists of two alternating looping stages: Period alignment and Pose optimization, as shown in Figure~\ref{fig:framework}.  The \textit{Period Alignment} stage identifies multiple reliable poses to calculate the average reference pose. The \textit{Pose Optimization} stage optimizes unreliable poses based on this reference pose and the VIO poses.
\sysname periodically goes back to the \textit{Period Alignment} phase to recalculate the reference pose and thus negate the effect of VIO drift (see Section~\ref{sub:DA}).


\textbf{Period Alignment}: The Period alignment stage checks the odometry of consecutive $N+1$ images. It is in a sequential localization mode since the system can only determine the absolute pose with at least $N+1$ images as input. We consider $I_j$ the first image to enter the period alignment stage. When all $N$ consecutive pairs of images satisfy the requirement that $\text{RPE}_{<apr,vio>} \leq d_{th}$ and  $\text{ROE}_{<apr,vio>} \leq o_{th}$ the predicted results $\{\hat{p}_i\}^{j+N+1}_{i=j}$ of these $N+1$ consecutive images from $I_j$ to $I_{j+N+1}$ can considered as accurate predicted outputs.  $d_{th}$ and $o_{th}$ are the relative pose checker's distance and orientation threshold to filter the inaccurate estimated poses.
    If the difference between $\hat{u}_{i,i+1}$ and $u_{i,i+1}^{vio}$ is larger than $d_{th}$ or $o_{th}$, the pose $\hat{p}_i$ is inaccurate and discarded.
    If the difference between $\hat{u}_{i,i+1}$ and $u_{i,i+1}^{vio}$ is less than $d_{th}$ and $o_{th}$ simultaneously,  poses $\hat{p}_i$ and $\hat{p}_{i+1}$ pass the relative pose checker.
    Upon getting $N+1$ reliable estimated poses,
     we perform geometric averaging from $\{\hat{p}_i\}^{j+N+1}_{i=j}$ using Weiszfeld's algorithm\footnote{\url{https://pypi.org/project/geom-median/}.} and ~\cite{gramkow2001averaging}\footnote{\url{https://github.com/christophhagen/averaging-quaternions}} to get reference pose $\Bar{P}^{apr}$ since we assume the pose error of APR is normally distributed in space.  We  perform the same geometric averaging from  $\{p_i^{vio}\}^{j+N+1}_{i=j}$ to get the reference pose $\Bar{P}^{vio}$ in the VIO system. Once $\Bar{P}^{apr}$ and $\Bar{P}^{vio}$ are defined, \sysname moves to the \textit{Pose Optimization} stage. 
     

 \textbf{Pose Optimization:}
The \textit{Pose Optimization} calculates the pose estimation for each query image.
    The Relative Pose Checker checks the predicted pose against the previous image's pose as follows. 
    For the  subsequent $T$ predicted poses $\{\hat{p}_i\}_{i = j+N+2}^{T + j+N+2}$ and their corresponding poses $\{p_i^{vio}\}_{i = j+N+2}^{T + j+N+2}$ in VIO system, we  get the odometry $<\hat{\Delta}_{trans},\hat{\Delta}_{rot}>$ of $\hat{p}_i$ and $\Bar{P}^{apr}$ in APR coordinates and odometry $<\Delta_{trans}^{vio},\Delta_{rot}^{vio}>$ of $p_i^{vio}$ and $\Bar{P}^{vio}$ in VIO coordinates respectively.  If $\text{RPE}_{<apr,vio>} \leq d_{th}$ and  $\text{ROE}_{<apr,vio>} \leq o_{th}$ is satisfied, we consider the APR output for this image to be accurate as the difference of odometry between APR and VIO is very small. The pose is considered reliable and can be output directly. Otherwise, the pose is optimized by $OptimizePose(p^{vio},\Bar{P}^{apr},\Bar{P}^{vio})$ (see Algorithm~\ref{alg:opt}).
We calculate the rigid transformation between VIO coordinates and world coordinates using $\Bar{P}^{apr}$ and $\Bar{P}^{vio}$. Then we transform $p^{vio}$ to world coordinates and replace the unreliable pose.
    Once the system reaches frame $T+1$, it loops back to the \textit{Period Alignment} stage to find a new reference pose and compensate for VIO drift.

\begin{algorithm}
\caption{OptimizePose($p^{vio},\Bar{P}^{apr},\Bar{P}^{vio}$)}
\label{alg:opt}
\begin{algorithmic}[1]
\Require  $p^{vio} = <{\mathbf{x}^{vio}},{\mathbf{q}^{vio}}>$,$\Bar{P}^{apr} = <{\mathbf{\Bar{x}}}^{apr},{\mathbf{\Bar{q}}}^{apr}>$,$\Bar{P}^{vio} = <{\mathbf{\Bar{x}}}^{vio},{\mathbf{\Bar{q}}}^{vio}>$ .

\State $\mathbf{q}_{rel} \gets \mathbf{\Bar{q}}^{apr-1}\mathbf{\Bar{q}}^{vio}$
\State $\mathbf{R}_{rel} \gets QuaternionToRotation(\mathbf{q}_{rel})$
\State $\mathbf{T}_{rel} \gets \mathbf{\Bar{x}}^{apr} -\mathbf{R}_{rel} (\mathbf{\Bar{x}}^{vio})^T$
\State $\mathbf{x}_{opt} \gets \mathbf{R}_{rel} (\mathbf{x}^{vio})^T + \mathbf{T}_{rel}$
\State $\mathbf{q}_{opt} \gets \mathbf{q}^{vio} \mathbf{q}_{rel}^{-1} $
\State \textbf{Return} $<\mathbf{x}_{opt},\mathbf{q}_{opt}>$
\end{algorithmic}
\end{algorithm}

As shown in Figure~\ref{fig:framework}, \textbf{we call the $N+1$ consecutive query images of reliable poses in \textit{Period Alignment} stage and query images pass the relative pose checker in \textit{Pose Optimization} stage as keyframes.}
  Such a framework removes some estimated poses with particularly large errors in the \textit{Period Alignment} stage. Predicted poses of keyframes are considered to be reliable and close to the GT. 
  The average estimated pose of keyframes in the Period alignment stage and the corresponding VIO poses are reference poses, $\Bar{P}^{apr}$ and $\Bar{P}^{vio}$. Then, we correct unreliable poses and keep reliable poses in the \textit{Pose Optimization} stage according to $\Bar{P}^{apr}$ and $\Bar{P}^{vio}$. We show that the \sysname framework greatly improves the accuracy of the median error and reduces the output of poses with large errors by identifying keyframes and optimizing partial poses in Section~\ref{sec:exp}.

\section{Dataset}
\label{sec:ds}
We introduce a localization dataset for this paper, consisting of 70 recordings captured in six locations, including three outdoor and three indoor scenes. 
We develop a simple application integrates ARKit 6\footnote{\url{https://developer.apple.com/augmented-reality/}}, Gyroscope API\footnote{\url{https://docs.unity3d.com/ScriptReference/Gyroscope.html}}, and iOS GPS Plugin\footnote{\url{https://assetstore.unity.com/packages/tools/localization/native-gps-plugin=ios-android-216027\#content}} based on Unity Engine to collect VIO system's poses data, IMU data, and GNSS location data. All data and timestamps were recorded simultaneously with the image capture. The resolution of all images is $1920\times1440$. All images are fed into an SfM framework using COLMAP~\cite{schonberger2016structure} to get the GT.  Participants who recorded sequences were asked to freely walk through each scene with the phone in hand without specific routes. Figure~\ref{fig:outhis} shows the speed histogram when collecting data. 
For outdoor scenes, we captured images in 1Hz. For indoor scenes, we captured images in 2Hz.  All visible faces and license plates are anonymized. Table~\ref{tab:ds} provide summary statistics on the dataset, including the number of samples in the testing and training sets, both derived from independent sequences. 

 \begin{table}[t]
\caption{Dataset details.}
$$
\begin{array}{c|c|cc|c} 
& \text{Scenes} & \multicolumn{2}{c|}{\text { Dataset quantity }} & \text { Spatial } \\
&\text {} & \text { Train } & \text { Test } & \text { Extent }(\mathrm{m}) \\
\hline 
&\text { Square } & 2058 & 1023&  40\times25 \\
\text{Outdoor}&\text { Church } & 1643 & 853  &  50\times40 \\
&\text { Bar } & 1834 & 838 & 55\times 35 \\
\hline 
&\text { Stairs } &  873& 222 &  5.5 \times 4.5 \times 6 \\
\text{Indoor}&\text {Office } &  1479 &635  & 7.5 \times 4 \\
&\text { Atrium } & 1694 & 441 &  30 \times 50\\
\hline 
\end{array}
$$
\label{tab:ds}
\end{table}

\subsection{Outdoor}
Our dataset contains three large locations representative of AR use cases: 1) Square ($1000m^2$) is an urban scene at an intersection that contains several office buildings and a garden. 2) Bar ($1925m^2$) is a scene comprising three historic buildings and outdoor bars. 3) Church ($2000m^2$) is a scene composed of a historical building and a small garden. 1) and 2) are typical scenes in modern cities, with abundant moving objects like pedestrians and vehicles. 3) is more like a landmark building in a tourist attraction. In addition, each dataset was collected over multiple days and periods, incorporating changes in weather and light as well as changes in the environment's appearance. Figure~\ref{fig:imageexample}.a gives example frames from each scene.

\subsection{Indoor}
\begin{figure}[!t]
 \centering
  \subfigure[Images for outdoor scenes]{
 \label{fig:subfig:a} 
 \includegraphics[width=0.9\linewidth]{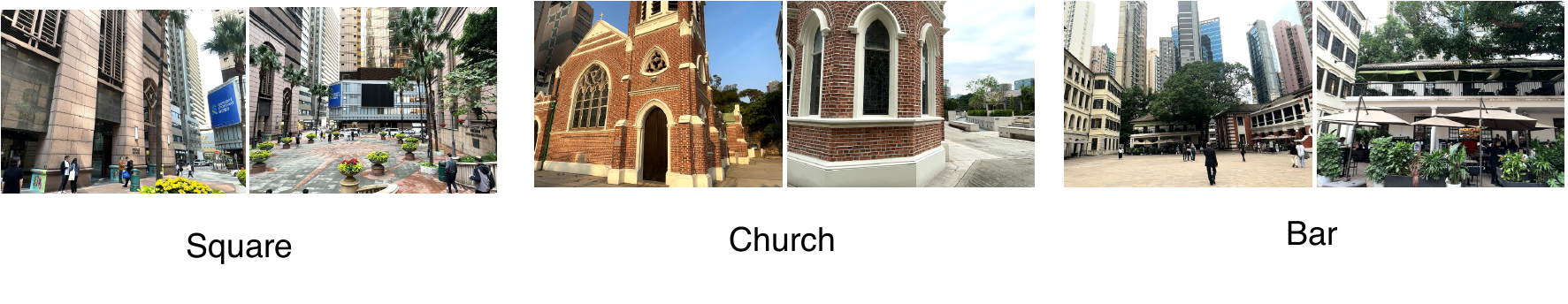}}
 \subfigure[Images for indoor scenes]{
 \label{fig:subfig:b} 
 \includegraphics[width=0.9\linewidth]{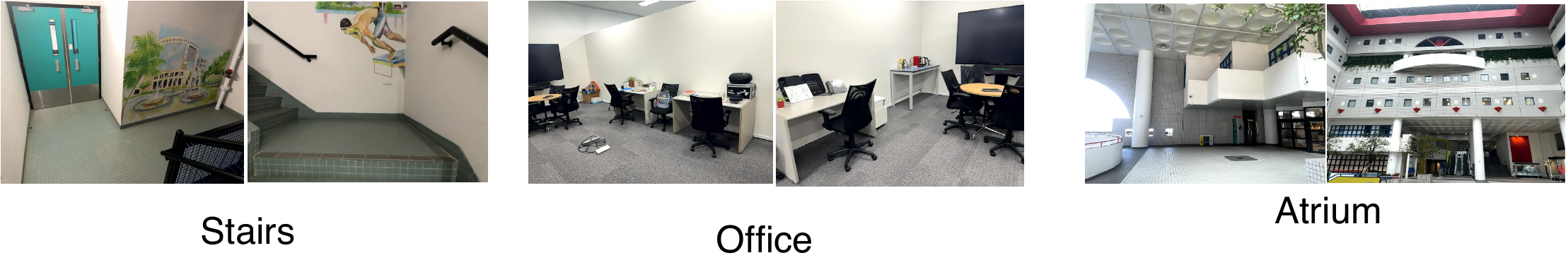}}
 \caption{Example frames from the (a) outdoor and (b) indoor datasets. There are 40 sequences for 3 outdoor scenes, each containing around 200 images; there are 30 sequences for 3 indoor scenes, each containing between 100 (Stairs) and 200 (Office, Atrium) images. }
 \label{fig:imageexample} 
\end{figure}

The initial release of our dataset also contains 3 indoor locations representative of
AR use cases: 1) Stairs ($149 m^3$) is the staircase between two floors of a building with murals on the walls.  2) Office ($30m^2$) is a daily office and meeting scenario. 1) and 2) are very similar to previous indoor visual localization data sets~\cite{glocker2013real,shotton2013scene}.
3) Atrium ($1500 m^2$) is a large indoor scene with many moving people of a local university.  It was collected over multiple days and multiple time periods. Besides, the scene has small changes in the appearance for different days. Figure~\ref{fig:imageexample}.b gives example frames from each scene.

\begin{figure}[!t]
 \centering
  \subfigure[Outdoor]{
 \label{fig:subfig:f} 
 \includegraphics[height=1.0in,width=1.4in]{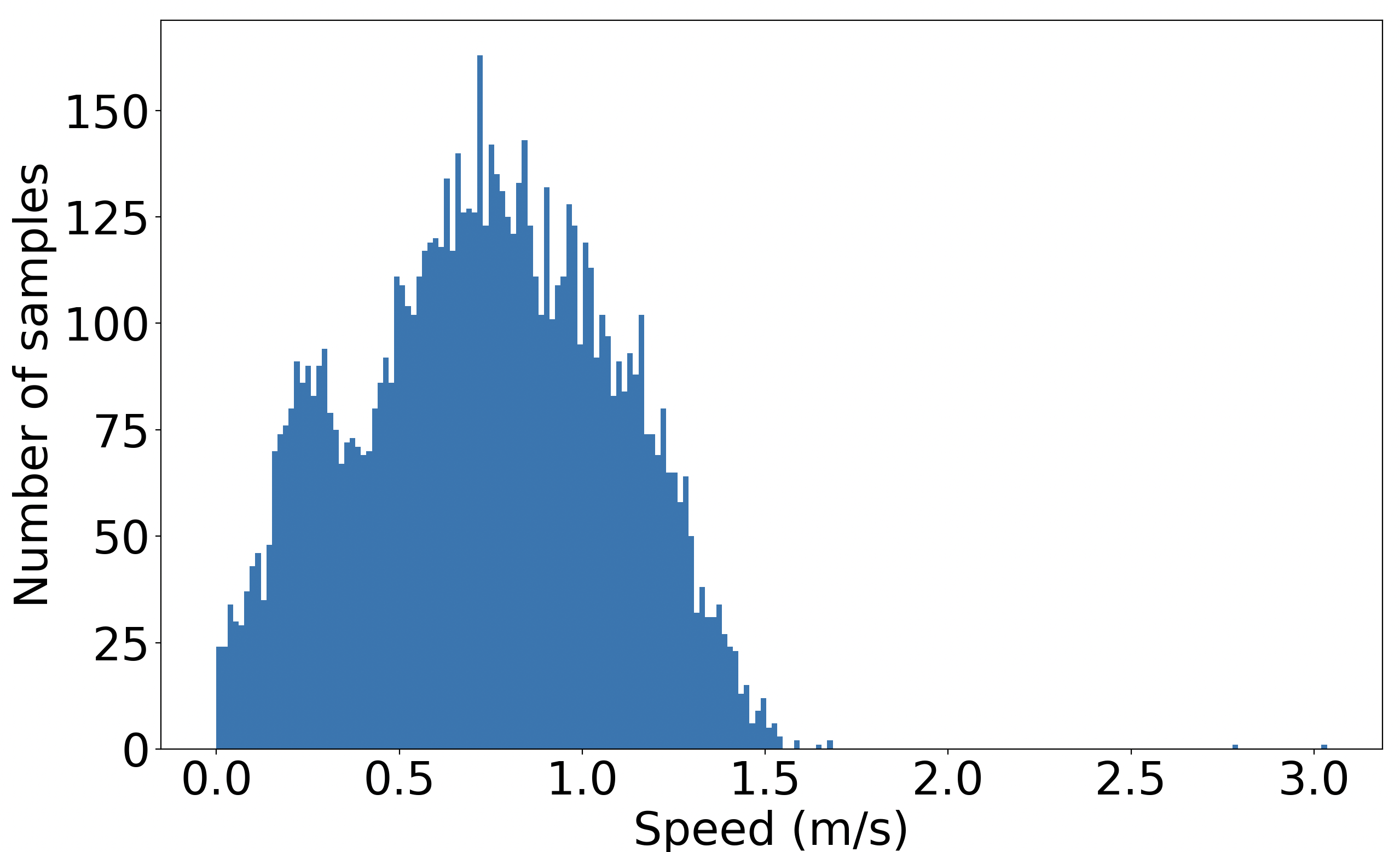}}
 \subfigure[Indoor]{
 \label{fig:subfig:a} 
 \includegraphics[height=1.0in,width=1.4in]{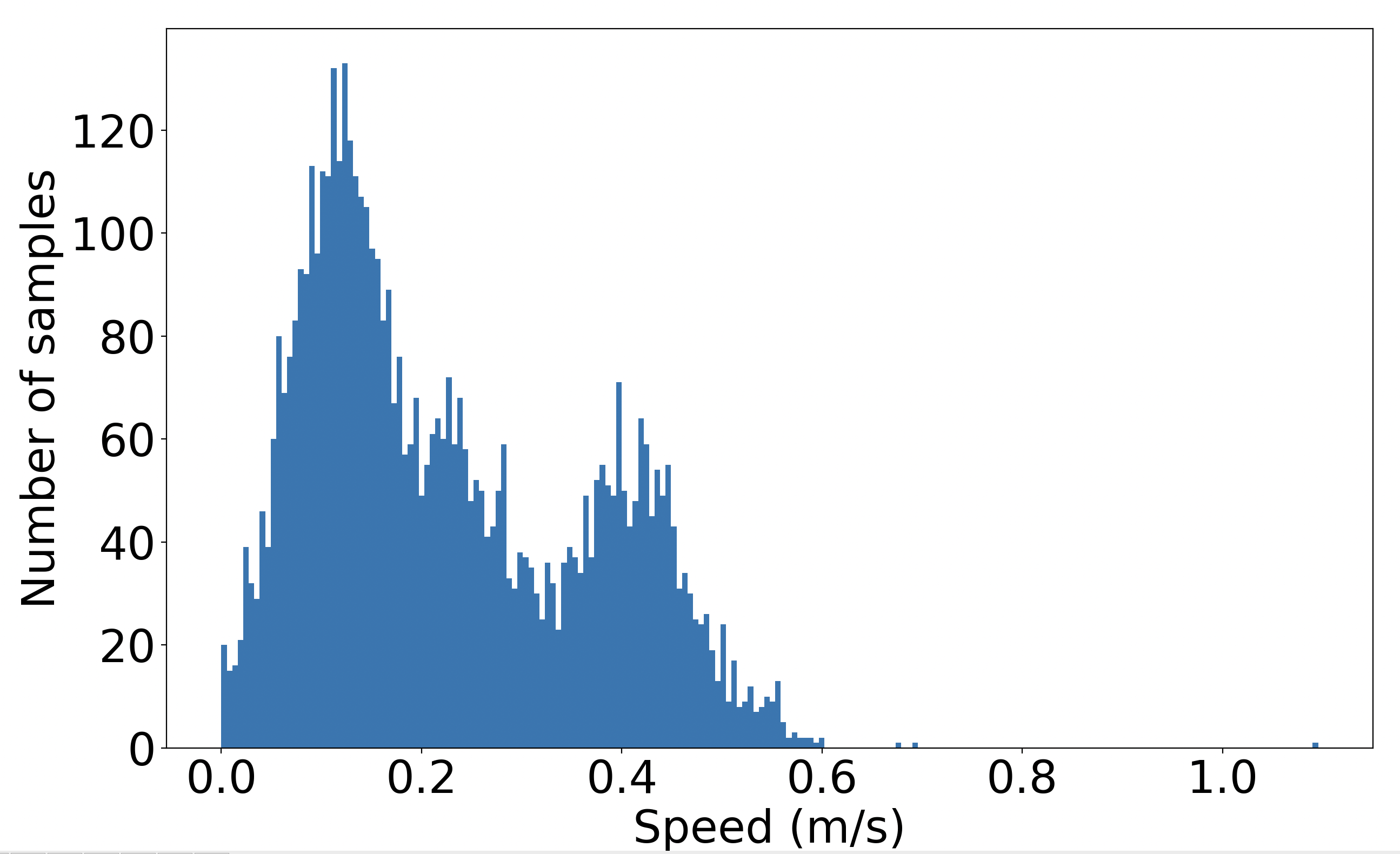}}
 \caption{Speed histograms for all data captures in the outdoor (a) and indoor (b) datasets.}
 \label{fig:outhis} 
\end{figure}

\section{Implementation}

\begin{figure}[!t]
 \centering
  \subfigure[Stairs]{
 \label{fig:subfig:a} 
 \includegraphics[height=0.8in,width=1.2in]{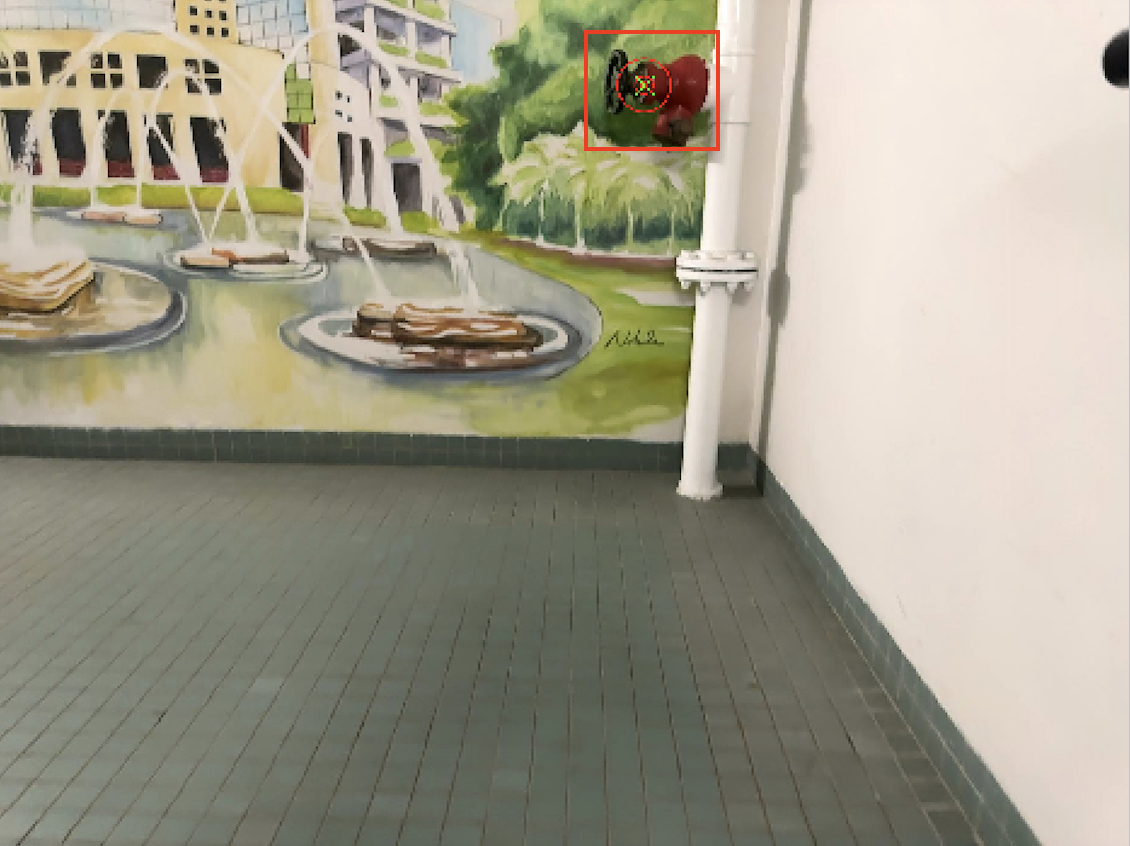}}
 \subfigure[Church]{
 \label{fig:subfig:b} 
 \includegraphics[height=0.8in,width=1.2in]{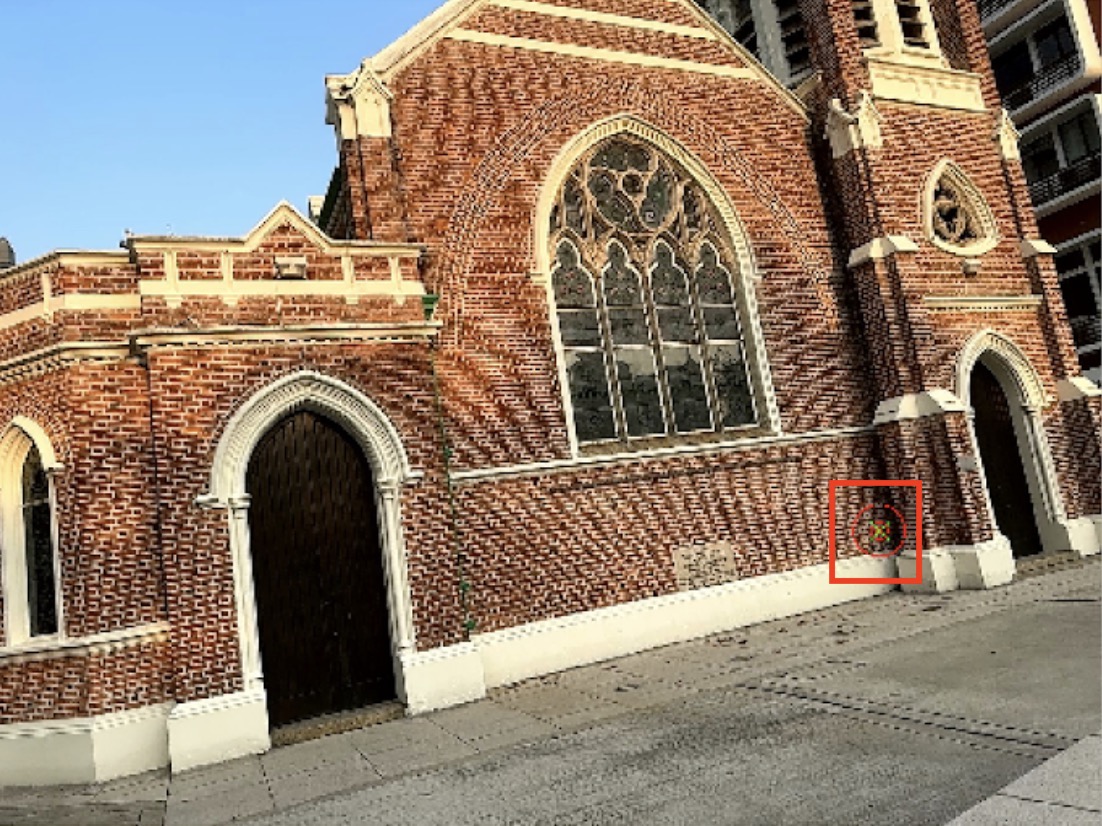}}
 \caption{GT positions of the bunny on two of the collected datasets to compare the AR effect of our method with original APR methods.}
 \label{fig:bunnygt} 
\end{figure}

We implement \sysname both on a desktop computer to conduct experiments on the collected dataset and in a mobile AR app for collecting system performance data.

\subsection{Desktop Implementation}
\label{subsec:imp}

There are four hyperparameters. $d_{th}$ and $o_{th}$ are the Relative pose checker's distance and orientation thresholds to filter inaccurate pose estimations.  
If among $N+1$ consecutive images, $N$ image pairs pass the relative pose checker, we consider these pose estimations to be accurate. $T$ is the number of images processed in Pose Optimization before returning to the Period Alignment. We set $d_{th} = 0.4m$ and $o_{th} = 4^ \circ$ according to the average RPE$_{<vio,GT>}$ and ROE$_{<vio,GT>}$ over our datasets. We set $N = 2$ and $T = 8$, with one frame processed per second for the AR application demo and all the experiments in this paper. We implement our framework over two APR models:

\noindent\textbf{PN}. PoseNet (PN) is the baseline method in this paper. Since there is no open source code for PoseNet~\cite{kendall2015posenet,kendall2017geometric}, we  follow~\cite{brahmbhatt2018geometry,melekhov2017image} and use ResNet34 as the backbone network. 


\noindent\textbf{MS-T}. MS-Transformer~\cite{shavit2021learning} (MS-T) is the SOTA model in supervised learning APR. It aggregates the activation maps with self-attention and queries the scene-specific information. MS-T extends the single-scene paradigm of APR to learning multiple scenes in parallel. Therefore, we trained one MS-T for all outdoor scenes and one MS-T for all indoor scenes using the official open-source code\footnote{\url{https://github.com/yolish/multi-scene-pose-transformer}}.

\textbf{We note APR methods integrated into our \sysname framework as $\text{APR}^{vio}$}.
All APR models are implemented in PyTorch~\cite{paszke2019pytorch}. During training, all input images are resized to $256\times 256$ and then randomly cropped to $224\times 224$ before. For both PoseNet and MS-T, we set an initial learning rate of $\lambda= 10^{-4}$. For all APR models, we use a batch size of 32. The training epoch of PN is 400 for all datasets. The training epoch of MS-T is 600 for all dataset. We implement our framework as a Python script set that takes the output pose from the selected APR method and applies the algorithms discussed in Section~\ref{sec:method}.  All experiments for evaluation in Section~\ref{sec:exp} are performed on an NVIDIA GeForce GTX 3090 GPU.

\subsection{Mobile AR Demo}
\label{subsec:mobilear}
Besides the Python scripts described above, we implement \sysname as a mobile AR app using Unity and ARKit to run on an  iPhone 14 Pro Max.  We converted the pre-trained PN with a ResNet34 backbone to ONNX format and incorporated it into a Unity application. We use OpenCVforUnity\footnote{\url{https://assetstore.unity.com/packages/tools/integration/opencv-for-unity-21088}} for processing query images and use Barracuda transfering resized images to tensor as the input of PN. We deploy the framework described in Section~\ref{subsec:vio-apr} as a set of C\# scripts.

\begin{figure}[!t]
\centering
  \subfigure[PoseNet (Stairs)]{
 \label{fig:subfig:a} 
 \includegraphics[width=0.9\linewidth]{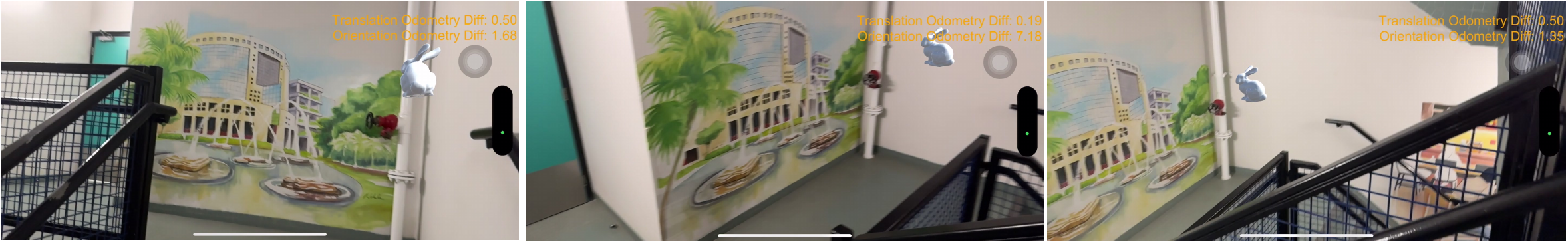}}
 \subfigure[PoseNet$^{vio}$ (Stairs)]{
 \label{fig:subfig:b} 
 \includegraphics[width=0.9\linewidth]{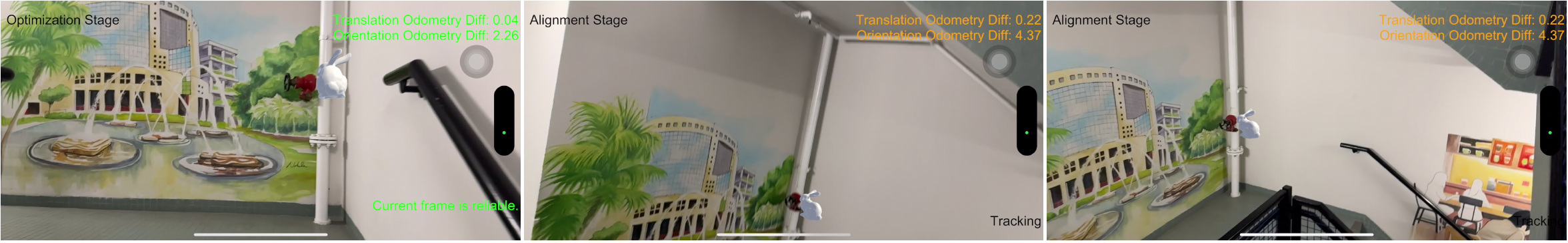}}
  \subfigure[PoseNet (Church)]{
 \label{fig:subfig:a} 
 \includegraphics[width=0.9\linewidth]{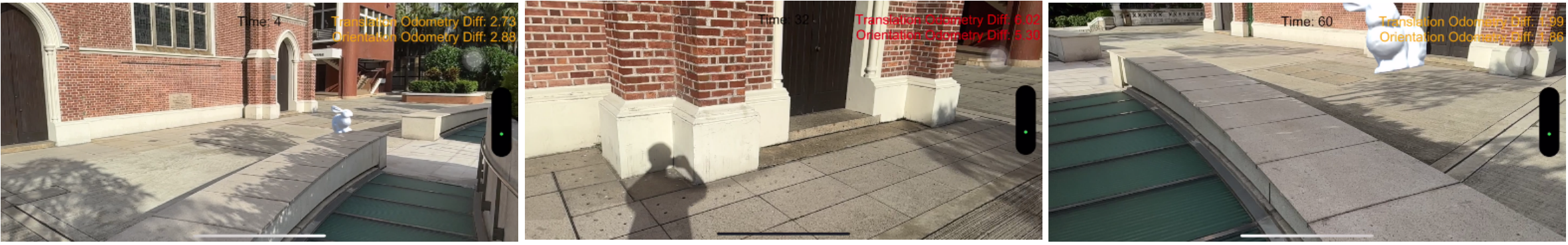}}
 \subfigure[PoseNet$^{vio}$ (Church)]{
 \label{fig:subfig:b} 
 \includegraphics[width=0.9\linewidth]{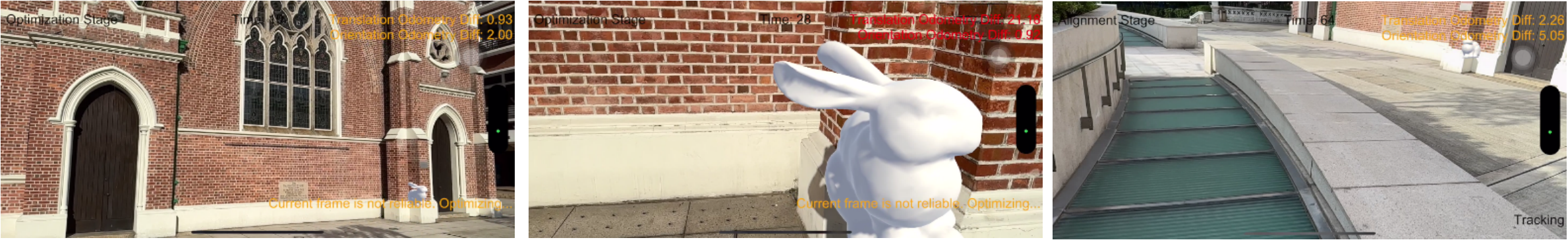}}
 \caption{AR frames displaying the 3D bunny at the GT pose (see Figure.~\ref{fig:bunnygt}) using PoseNet and PoseNet$^{vio}$ (ours). 
 PoseNet$^{vio}$ keeps the bunny close to the GT with less variability than PoseNet alone.
 }
 \label{fig:areffec} 
\end{figure}

In mobile marker-less AR applications, assume we need to place some virtual objects in specific coordinates in SfM coordinates of the scene. 
Once reliable absolute poses are obtained, we can adjust (1) the AR Session Origin's pose~\footnote{\url{https://docs.unity3d.com/Packages/com.unity.xr.arfoundation@1.0-preview.8/api/UnityEngine.XR.ARFoundation.ARSessionOrigin.html}} to place the AR Camera at the original point and the desired orientation and (2) the object's coordinate based on the position of the output pose. With the adjustment, we are able to render virtual objects correctly. The local tracking module in ARkit or ARCore then automatically keeps the place of virtual objects during the empty window period ($I^*, I_{t-2}, I_{t-1}, I_{t}, I_{t+1}$) as shown by the green arrows in Figure~\ref{fig:framework} and Figure~\ref{fig:drift}. In practice, we use Camorph~\cite{Brand2022CAMORPHAT} for alignment because COLMAP and Unity have coordinate systems with different orientation of axis and handedness. As shown in the Figure~\ref{fig:bunnygt}, we select two coordinates for placing an open-source 3D object, the Stanford Bunny, in the COLMAP reconstructed models of Stairs and Church, respectively. The position where the bunny is located should be stable and the same as the position in SfM coordinates. As depicted in Figure~\ref{fig:areffec}, the localization error of PN is very large under some query images.
Instead, the localization result of our method is in a small range because our method does not trust all APR predictions, but find the reliable APR output by comparing the difference between the odometry of APR predictions and the odometry of VIO between consecutive images until it finds the reliable APR output.   Since our method filtered unrelible predicted poses and improves the robustness of the estimation, the movement of the bunny is small between each update. Local tracking usually occurs only as a slow drift unlike the output of APR jumps between even adjacent frames.  Please refer to the supplementary video for watching the AR effect.

\subsection{Design Analysis}
\label{sub:DA}

\begin{figure}[t]
  \centering
  \includegraphics[width=.8\linewidth]{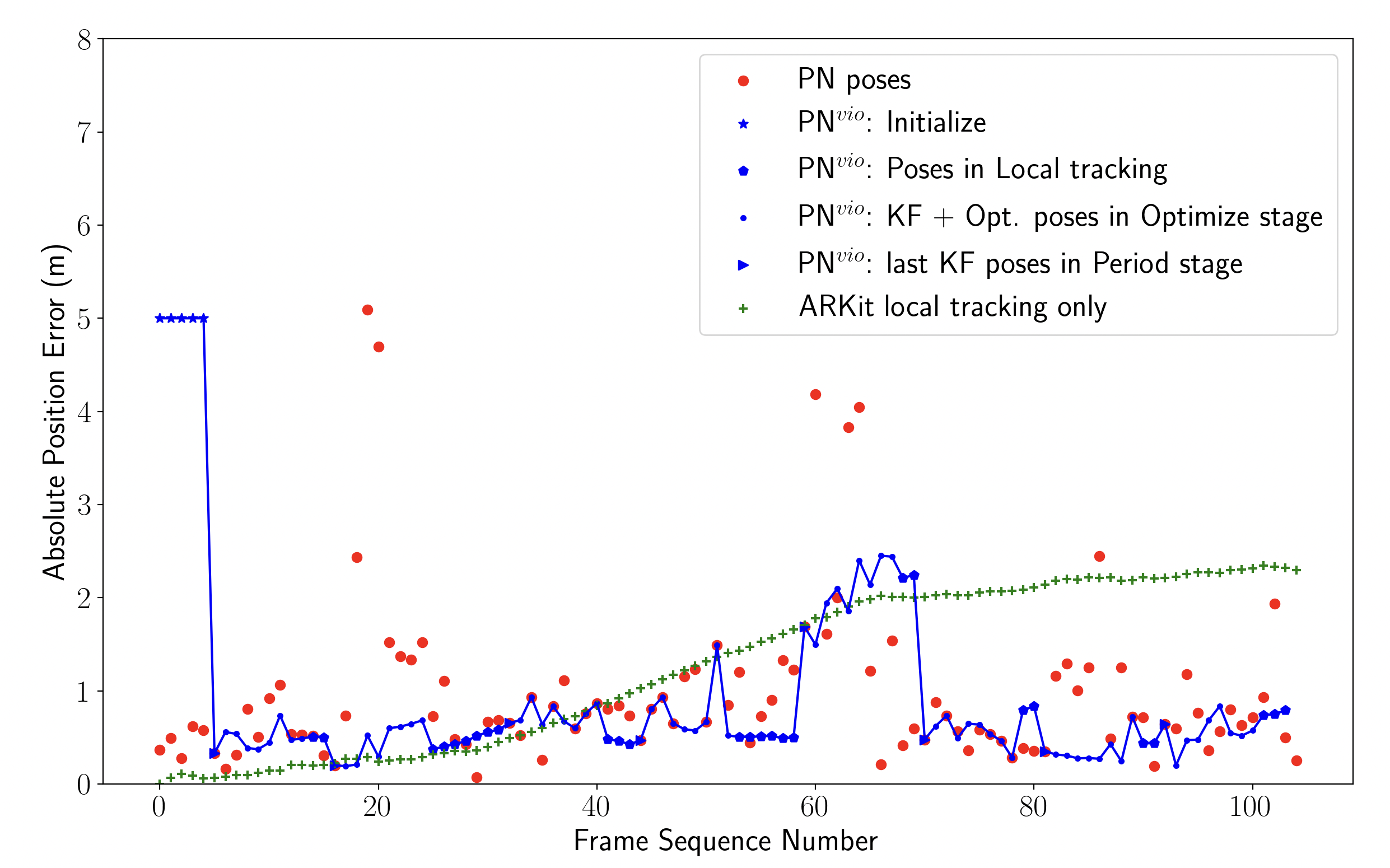}
\caption{Position error for PN (red), VIO only (green), and each stage of PN$^{vio}$ (blue) on one test sequence of the Church dataset. The VIO shows noticeable drift over time, while PoseNet displays highly random inaccuracy. PN$^{vio}$  keeps the pose error small in the optimization stage and reduces the VIO drift during period alignment. 
}
\label{fig:drift}
\end{figure}

Using ARKit's local tracking module %
for bridging between trusted pose estimates is effective. However, drift accumulates with motion. 
Since each image of our dataset has ARKit pose labels, we assume that the rigid transformation between the SfM coordinates and ARKit coordinates of the first image's GT of a sequence is known, and then we use the tracking module for absolute pose estimation, as the pure local tracking error shown in the Figure~\ref{fig:drift}, the error gradually becomes larger with motion, and then stabilize above two meters. This is why we put a restriction $T$ on the optimize stage so that after getting the reference pose, we need to get a new reference pose every once in a while by period alignment. This ensures that our method can get much more stable and much lower error than pure local tracking and PoseNet without knowing any query's GT.

\section{Experiment}

\begin{figure}[!t]
 \centering
  \subfigure[Position error distribution]{
 \label{fig:subfig:a} 
 \includegraphics[height=1.0in,width=1.5in]{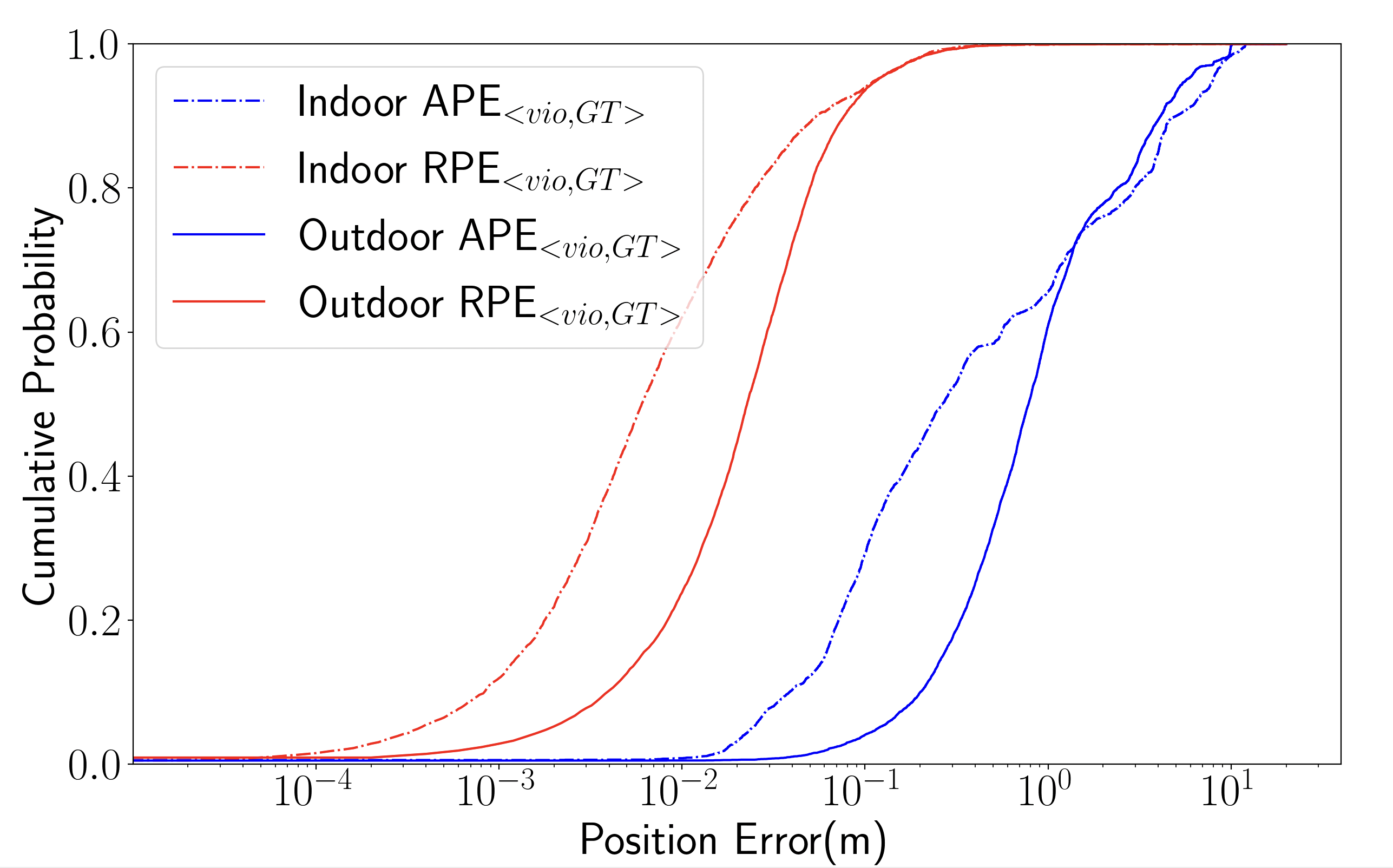}}
 \subfigure[Orientation error distribution]{
 \label{fig:subfig:b} 
 \includegraphics[height=1.0in,width=1.5in]{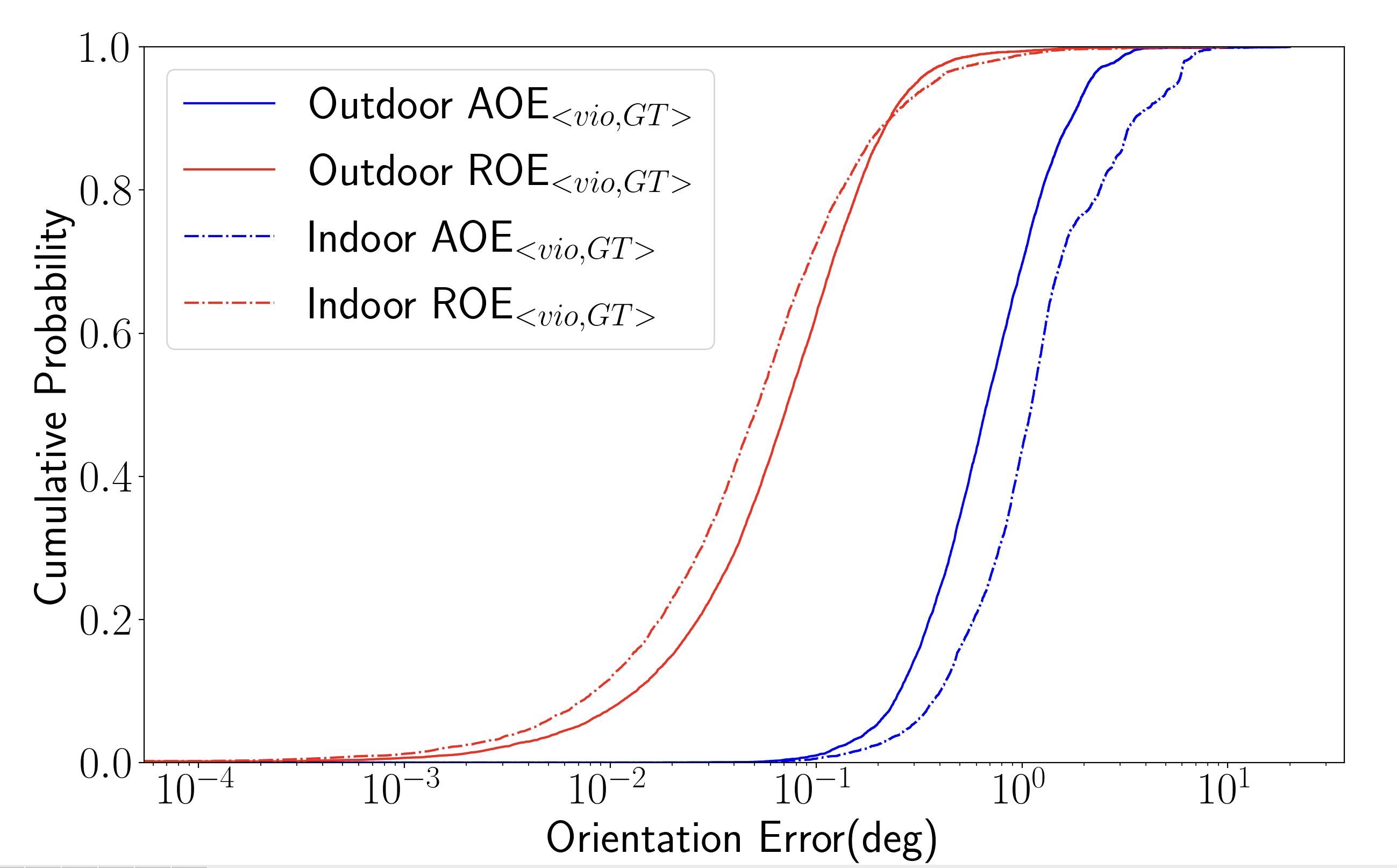}}
 \caption{ (a) Cumulative distribution of ARkit's position error. 
 (b) Cumulative distribution of ARkit's orientation error.}
 \label{fig:errdis} 
\end{figure}

\label{sec:exp}
Since there is no well-known dataset for AR scenario where each image has both a GT pose and pose label from the mobile VIO system, we evaluate our framework on the datasets described in Section~\ref{sec:ds}.  To calculate the absolute pose error and the  relative pose error between the GT and the ARkit pose, we estimate a rigid transform (3D rotation and translation) between the ARKit tracks and the GT based on the first 30s of estimates in each
method (each sequence is around 200s)~\footnote{For the smallest stairs scene, the length of each sequence is around 100s}. The aligned tracks all start from the origin. We follow the Equation~(\ref{eq:cum})~\cite{cortes2018advio}, as well as Equations~(\ref{eq:ape}, \ref{eq:aoe}, \ref{eq:rpe}, \ref{eq:roe}) to obtain the cumulative distribution function for the absolute pose error (APE/AOE) and the relative pose error (RPE/ROE):
\begin{equation}
\hat{F}_n(d)=\frac{\text { number of pose errors } \leq d}{n}
\label{eq:cum}
\end{equation}

\begin{equation}
    \text{APE}_{<vio, GT>}  = || \mathbf{x}_i - \mathbf{x}^{vio}_i||_2
\label{eq:ape}
\end{equation}

\begin{equation}
    \text{AOE}_{<vio, GT>}  = 2\arccos{|\mathbf{q}_{i}^{-1}\mathbf{q}_{i}^{vio}|}\frac{180}{\pi}
\label{eq:aoe}
\end{equation}
\noindent where $n$ is the number of poses. $\mathbf{x}_i$ is GT, and $\mathbf{x}_i^{vio}$ is the label from the VIO system. In our dataset, $\mathbf{x}_i^{vio}$ are read from Apple ARKit. 
Figure~\ref{fig:errdis} illustrates the cumulative distribution of position (a) and orientation (b) error. Over 90\% of ARKit poses present less than 0.1m and 1° error, confirming  the  assumption that ARKit pose can generally be considered as ground truth. 

\begin{table}[t]
\caption{Median absolute position/orientation errors in m/° in outdoor scenes. ($N = 2, T = 8$)}
$$
\begin{array}{l|cc|cc}
\hline
\multicolumn{5}{c}{\text{Only Keyframes}}\\
\hline
\text{Outdoor} & \text {PN} & \text {PN}^{vio}\text{(ours)}  & \text {MS-T} & \text {MS-T}^{vio}\text{(ours)}  \\ \hline 
\text{Square}   &1.11 / 3.61&0.68/2.75&1.50/2.14&0.78/1.74\\
\text{Church}& 0.73 / 3.97&0.52/2.68&0.71/3.08&0.41/1.95\\
\text{Bar}  & 0.66 / 2.82&0.49/2.33&0.69/1.65&0.46/1.47\\ \hline
\text{average} & 0.83/3.47 & 0.56/2.59 &0.97/2.29 &0.55/1.72\\ 
\hline
\multicolumn{5}{c}{\text{Only Optimization}}\\
\hline
\text{Square}   & 1.08 / 3.69 &  0.77/2.64 & 1.55/2.31& 0.99/1.56\\
\text{Church} & 0.85 / 4.58 &  0.56/2.84 &0.74/2.79&0.59/1.63 \\
\text{Bar}  & 0.72/ 3.13  & 0.56/2.42   &0.78/1.79&0.58/1.32\\
\hline
 \text{average} &0.88/3.80 & 0.63/2.63 & 1.02/2.30 & 0.72/1.50 \\
\hline
\multicolumn{5}{c}{\text{Keyframes + Optimization }}\\

\hline
\text{Square}   &1.11 / 3.61 & 0.72/2.71 &1.50/2.14& 0.88/1.65\\
\text{Church} & 0.73 / 3.97 & 0.54/2.74 &0.71/3.08&0.47/1.80\\
\text{Bar}  & 0.66 / 2.82  & 0.52/2.39& 0.69/1.65&  0.51/1.43\\
 \hline
\text{average} &0.83/3.47 & 0.59/2.61  &0.97/2.29 & 0.62/1.63  \\\hline
\end{array}
$$
\label{tab:ape_kopt_out}
\end{table}

\begin{table}[t]
\caption{Median absolute position/orientation errors in m/° in indoor scenes. ($N = 2, T = 8$)}
$$
\begin{array}{l|cc|cc}
\hline
\multicolumn{5}{c}{\text{Only Keyframes }}\\
\hline
 \text{Indoor} & \text {PN} & \text {PN}^{vio}\text{(ours)}  & \text {MS-T} & \text {MS-T}^{vio}\text{(ours)}  \\ \hline 
 \text{Stairs} & 0.27/4.89&0.21/4.45&0.18/4.33&0.15/3.42\\
\text{Office}& 0.36/6.21 &0.30/4.93&0.35/4.66& 0.30/3.70\\
\text{Atrium} &1.78/5.54&1.16/4.24&2.0/3.98&1.23/3.03\\
 \hline
 \text{average} &0.80/5.55 & 0.56/4.54 & 0.84/4.32& 0.56/3.38 \\
 \hline
\multicolumn{5}{c}{\text{Only Optimization }}\\
\hline
\text{Stairs}  &0.34/5.66&0.21/4.45&0.26/7.14&0.16/2.57\\
\text{Office} &0.44/7.67&0.38/7.72&0.44/6.22&0.39/5.12 \\
\text{Atrium}  &1.65/4.74&1.44/4.50 &1.47/3.98&1.77/3.18 \\
 \hline
\text{average} & 0.81/6.02 & 0.68/5.56  & 0.72/5.78 &  0.77/3.62\\
\hline
\multicolumn{5}{c}{\text{Keyframes + Optimization }}\\
\hline
\text{Stairs}  & 0.27/4.89  &  0.22/4.45&0.18/4.33&0.16/3.23\\
\text{Office} & 0.36/6.21  & 0.31/5.58&0.35/4.66& 0.31/3.96\\
\text{Atrium} &1.78/5.54  &  1.22/4.27 &2.0/3.98&1.51/3.12\\
 \hline
\text{average} & 0.80/5.55 & 0.58/4.77  & 0.84/4.32 & 0.66/3.44 \\\hline
\end{array}
$$
\label{tab:ape_kopt_in}
\end{table}

  We evaluate the performance of APR and our framework through two primary metrics. As indoor and outdoor scenes present significant differences in terms of scale, number of images, and capture speed (see Figure~\ref{fig:outhis} and Table~\ref{tab:ds}), we present the results separately. We consider the median APE and AOE in Tables~\ref{tab:ape_kopt_out} (outdoor scenes) and~\ref{tab:ape_kopt_in} (indoor scenes), including $\text{APE}_{<apr, GT>}$ and $\text{AOE}_{<apr, GT>}$, as shown in Equation (\ref{eq:ape_apr}) and Equation (\ref{eq:aoe_apr}) for all test frames. We also evaluate the percentage of test images with pose predicted with high ($0.25m, 2^\circ$), medium ($0.5m, 5^\circ$), and low ($5m, 10^\circ$) precision levels proposed by~\cite{sattler2018benchmarking} in Table~\ref{tab:ratio_outdoor} (outdoor scenes) and Table~\ref{tab:ratio_indoor} (indoor scenes). The higher the percentage of each precision level, the better the performance.

 \begin{equation}
    \text{APE}_{<apr, GT>}  = || \mathbf{\hat{x}}_i - \mathbf{x}_i||_2
\label{eq:ape_apr}
\end{equation}

\begin{equation}
    \text{AOE}_{<apr, GT>}  = 2\arccos{|\mathbf{q}_{i}^{-1}\mathbf{\hat{q}}_{i}|}\frac{180}{\pi}
\label{eq:aoe_apr}
\end{equation}

\begin{figure}[htb!]
 \centering
   \subfigure[Bar]{
   \centering
 \label{fig:subfig:f} 
 \includegraphics[width=.9\linewidth]{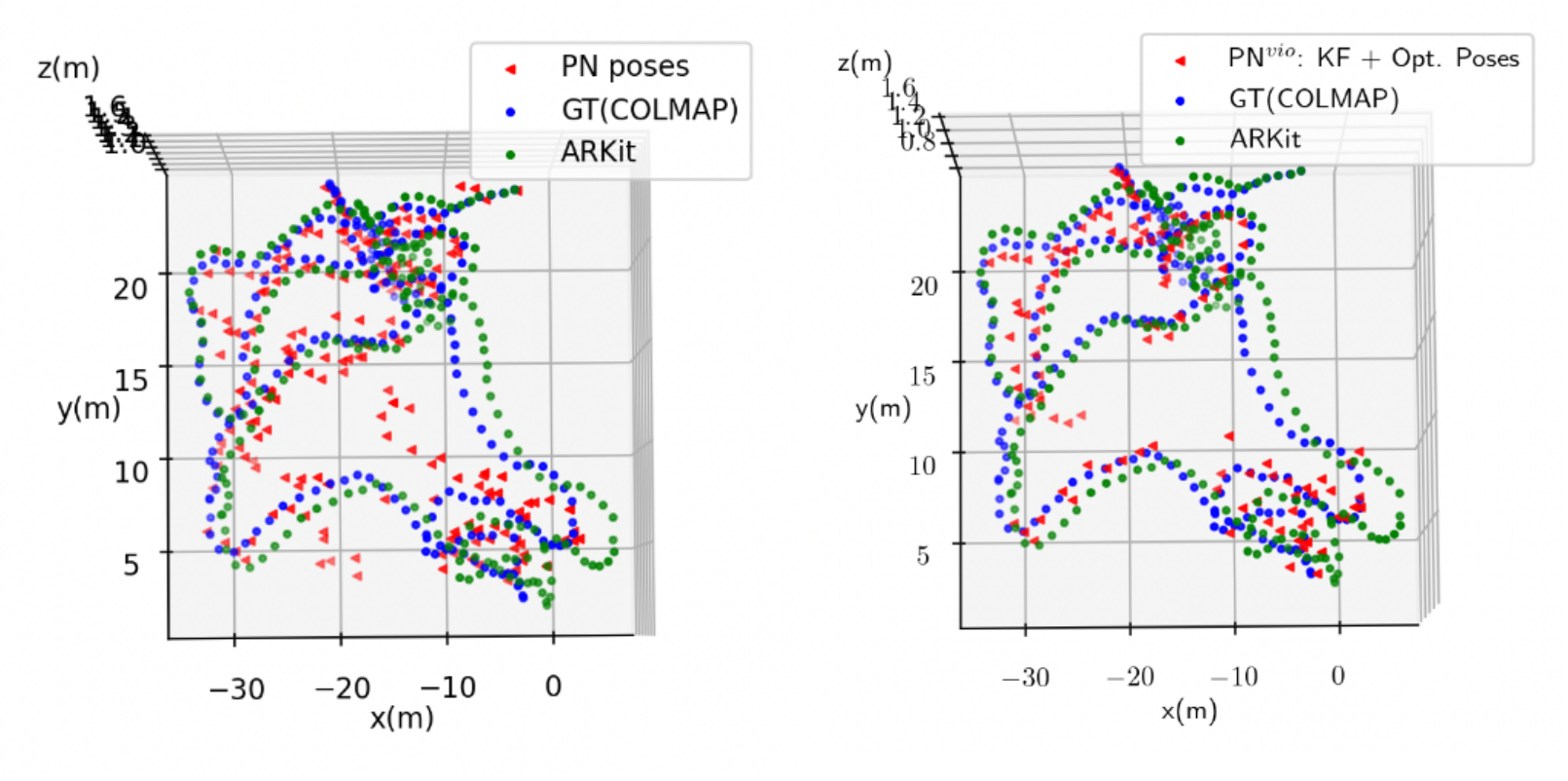}}
 \subfigure[Office]{
 \centering
 \label{fig:subfig:a} 
 \includegraphics[width=.8\linewidth]{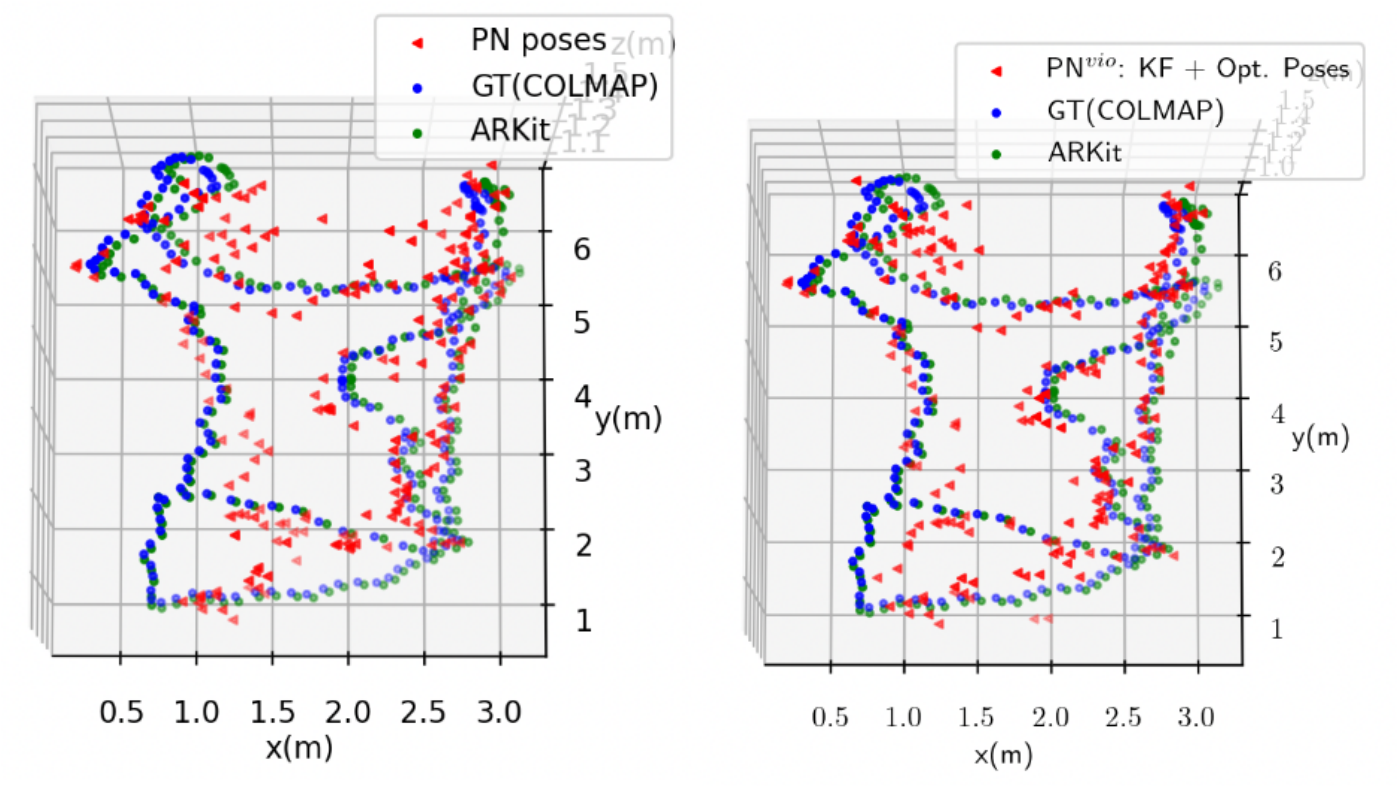}}
 \caption{Pose estimation for PN (left) and $\text{PN}^{vio}$ (right) for one test sequence in the Bar and Office scenes.
 Using \sysname significantly decreases the number of predictions with large error as well as the noisiness of pose estimation error compared to PN alone.
 }
 \label{fig:res_pn} 
\end{figure}

\subsection{Experiments on the outdoor dataset}

\noindent\textbf{Keyframes:} 
PN$^{vio}$ and MS-T$^{vio}$ significantly improve accuracy over PN and MS-T alone. Median APEs are greatly reduced as shown in Table~\ref{tab:ape_kopt_out}. In the three outdoor scenes, 
the keyframes preserved by PN$^{vio}$ reduce the position error by 25\% to 39\%. Orientation accuracy improved by 17\% to 32\%. The keyframes preserved by MS-T$^{vio}$ reduce the position error by 33\% to 48\%. Orientation accuracy improved by 21\% to 24\%. Table~\ref{tab:ratio_outdoor} shows that the percentage of each precision level is greatly increased by filtering out the unreliable poses that amount for 60 to 70\% of the total pose estimations. Pose estimates below the low accuracy level are greatly reduced in three scenes.
Almost no pose estimates of MS-T$^{vio}$ have more than 5 meters and 10 degrees pose error on the church scene. These results indicate that APR accuracy is partly pulled down by a significant portion of pose estimates that present a large error. Our pipeline uses VIO to identify and discard these pose estimates, resulting in  significantly higher accuracy.


\noindent\textbf{Optimized Poses:} Tables~\ref{tab:ape_kopt_out} and~\ref{tab:ratio_outdoor} confirm that 23\% to 33\% of the pose estimates that do not pass the relative pose checker in the pose optimization stage have larger error compared with the median pose error of all predictions.  
The percentage of almost all precision levels is increased by Algorithm~\ref{alg:opt} based on the reference poses from the Period alignment stage.  Pose estimates below the low accuracy level are greatly reduced for both PN$^{vio}$ and MS-T$^{vio}$. No pose estimates after optimizing of MS-T$^{vio}$ have more than 5 meters and 10 degrees pose error in the church scene. 
Optimizing unreliable poses may improve the median accuracy compared to filtering them out, while filtering keyframes may be beneficial in other scenarios. For instance, optimizing poses in MS-T$^{vio}$ results in higher orientation accuracy over all scenes compared to filtering keyframes at the cost of lower position accuracy. On the outdoor dataset, filtering keyframes and pose optimization thus address different issues with APR unreliable poses.


\noindent\textbf{Total results:} Combining keyframe selection with pose optimization leads to a much higher median accuracy than with keyframe selection or pose optimization alone, as shown in Tables~\ref{tab:ape_kopt_out} and~\ref{tab:ratio_outdoor}. MS-T$^{vio}$ increases the percentage of frames in the high
($0.25m, 2^\circ$) accuracy level by up to 112\% in Church. 
Figure~\ref{fig:res_pn} shows how PN$^{vio}$ improves accuracy. Compared to PN alone, PN$^{vio}$ reduces the incidence of outliers with large error, and reduces the noisiness of the prediction. It also reduces the impact of VIO drift.

\subsection{Experiments on the indoor dataset}

\begin{table*}[hbt!]
\caption{Percentage of keyframes' poses, optimized poses and total poses predicted with high (0.25m, $2^{\circ}$), medium (0.5m, $5^{\circ}$), and low (5m, $10^{\circ}$) accuracy~\cite{sattler2018benchmarking}. The value in parentheses represents the percentage of keyframes' poses, opimized poses, and total poses in the test set of outdoor scenes. }
$$
\begin{array}{c|l|cc|cc}
\hline
\multicolumn{6}{c}{\text{Only Keyframes}}\\
\hline
\text{Dataset} &\text{Scenes}  & \text {PN} & \text {PN}^{vio}\text{(ours)}   & \text {MS-T} & \text {MS-T}^{vio}\text{(ours)}  \\ \hline
 &\text{Square}  & 3.2/18.4/87.0 & 5.5/30.0/97.4 (30.3\%) &2.9/17.9/81.1& 3.3/27.3/93.6 (31.9\%)\\
\text{Outdoor} & \text{Church} & 2.2/21.9/82.1& 3.9/38.7/98.2 (32.7\%) &6.9/29.7/79.0&16.3/58.0/99.6(33.2\%)\\
&\text{Bar}  &  4.1/31.7/89.5 & 7.0/49.3/97.1 (40.8\%) &8.5/34.6/90.8 &14.8/53.1/99.4(40.2\%)\\
 \hline
\multicolumn{6}{c}{\text{Only Optimization}}\\
\hline
&\text{Square}  & 2.5/18.5/90.8 (30.7\%) & 1.3/28.3/95.5 (30.7\%)& 3.3/17.5/84.0 (32.9\%) & 6.2/19.9/92.3 (32.9\%)\\
\text{Outdoor}&\text{Church} & 1.9/14.9/80.4 (30.6\%) &  6.1/39.8/97.3 (30.6\%) &2.6/20.5/84.1(22.9\%) &12.3/39.0/100 (22.9\%)\\
&\text{Bar}  & 2.0/22.3/89.6 (30.0\%)   & 4.8/39.8/98.8 (30.0\%)  & 6.2/31.0/88.4(28.9\%) & 7.9/41.3/97.1(28.9\%)\\
 \hline
 \multicolumn{6}{c}{\text{Keyframes + Optimization}}\\
 \hline
 &\text{Square}  & 3.2/18.4/87.0 &  3.4/29.2/96.5(61.0\%) &2.9/17.9/81.1& 4.8/23.5/92.9 (64.8\%)\\
\text{Outdoor} & \text{Church} & 2.2/21.9/82.1 & 5.0/39.3/97.8(63.3\%)  &6.9/29.7/79.0& 14.6/50.2/99.8 (56.0\%)\\
&\text{Bar}  &  4.1/31.7/89.5 & 5.1/45.2/97.8 (70.8\%)  &8.5/34.6/90.8& 11.9/48.2/98.4 (69.1\%)\\
 \hline
\end{array}
$$
\label{tab:ratio_outdoor}
\end{table*}

 \begin{table*}[hbt!]
\caption{Percentage of keyframes' poses, optimized poses, total poses predicted with high (0.25m, $2^{\circ}$), medium (0.5m, $5^{\circ}$), and low (5m, $10^{\circ}$) accuracy~\cite{sattler2018benchmarking}. The value in parentheses represents the percentage of the keyframes' poses , opimized poses, and total poses in the test set of indoor scenes.}
$$
\begin{array}{cl|cc|cc}
\hline
\multicolumn{6}{c}{\text{Only Keyframes}}\\
\hline
\text{Dataset} &\text{Scenes}  & \text {PN} & \text {PN}^{vio}\text{(ours)}   & \text {MS-T} & \text {MS-T}^{vio}\text{(ours)}  \\ \hline
 &\text{Stairs}   & 5.4/48.2/87.4& 8.3/59.4/96.2 (59.9\%) &18.5/58.1/86.9&25/76.4/97.9 (64.9\%)\\
\text{Indoor}&\text{Office} & 3.8/31.0/72.8 & 6.4/42.2/86.6 (56.4\%) &7.1/46.0/80& 10.5/56.9/92.2 (62.8\%)\\
&\text{Atrium}  & 0.0/5.4/71.0 & 0.0/10.1/88.4(29.3\%)  &0.4/7.3/66.7&1.7/19.0/88.4 (27.4\%)\\
 \hline
 \multicolumn{6}{c}{\text{Only Optimization}}\\
 \hline
 &\text{Stairs}   & 1.9/38.5/80.8 (23.4\%)  &  3.8/63.4/100 (23.4\%)  &10.6/21.3/72.3 (21.2\%) &23.4/76.6/83.0 (21.2\%)\\
\text{Indoor}&\text{Office} &0/19.2/61.1 (32.0\%)&8.4/29.6/74.4 (32.0\%)& 1.6/31.7/68.9(28.8\%) &7.1/32.2/74.3 (28.8\%)\\
&\text{Atrium}  & 0/5.9/78.5 (30.6\%) & 0/3.0/80.7 (30.6\%)  &0/4.8/80.1(33.1\%) &2.1/10.3/85.6(33.1\%)\\
\hline
\multicolumn{6}{c}{\text{Keyframes + Optimization}}\\
\hline
 &\text{Stairs}   & 5.4/48.2/87.4 & 7.0/60.5/97.3 (83.3\%)  &18.5/58.1/86.9& 24.6/76.4/94.2(86.0\%)\\
\text{Indoor}&\text{Office} & 3.8/31.0/72.8  & 7.1/37.6/82.2 (88.3\%)  &7.1/46.0/80& 9.5/49.1/86.6 (91.7\%)\\
&\text{Atrium}  & 0/5.4/71.0 & 0/6.4/84.5 (59.9\%)  &0/7.3/66.7 & 1.9/14.2/86.9 (60.5\%)\\
 \hline
\end{array}
$$
\label{tab:ratio_indoor}
\end{table*}

\noindent\textbf{Keyframes:} Filtering out unreliable poses significantly improves median accuracy (see Table~\ref{tab:ape_kopt_in}). The keyframes preserved by PN$^{vio}$ reduce the position error by 17\% to 35\%. Median orientation accuracy improves by 9\% to 23\%. As shown in Table~\ref{tab:ratio_indoor}, pose estimates below the low accuracy level are significantly reduced in three scenes. Less than 4\% pose estimates of PN$^{vio}$ and MS-T$^{vio}$ have more than 5 meters and 10 degrees pose error for all indoor scenes. Despite a smaller-scale, constrained environment, some APR pose estimations still present a large error that lowers accuracy. Our method identifies these poses and retains those with a lower error.


\noindent\textbf{Optimized Poses:} Tables~\ref{tab:ape_kopt_in} and~\ref{tab:ratio_indoor} show that the  pose estimates that do not pass the relative pose checker in the pose optimization stage (21 to 33\% of all poses) have larger error than the median pose error. No pose estimates of PN$^{vio}$ have more than 5 meters and 10 degrees pose error on Stairs. The optimization on median pose error also has a bigger improvement than simply filtering out unreliable output.  The percentage of almost all precision levels is increased.

\noindent\textbf{Total results:} Tables~\ref{tab:ape_kopt_in} and~\ref{tab:ratio_indoor} demonstrate that keeping only the keyframes' estimated poses and optimizing some unreliable predicted poses achieves much higher median accuracy than all the original test frames, and the performance is improved at almost every precision level. Similar to the outdoor set, Figure~\ref{fig:res_pn} shows that  PN$^{vio}$ achieves higher accuracy by filtering out outliers, improving accuracy of PN poses, and reducing the impact of VIO drift.


\subsection{Analysis}
Table~\ref{tab:ratio_outdoor} shows that PN and MS-T have 10\% to 20\%  estimates that are very inaccurate in outdoor scenes. However, the probability of encountering large errors in three consecutive images while maintaining an odometry that closely matches VIO's is quite low. 
As such, \sysname uses the VIO system to identify the accurate APR pose estimates. These reliable poses are then used to calculate the reference pose.
Outdoor, 
our framework improves the accuracy of $\text{MS-T}$ by $36\%$ on median position error and $29\%$ on median orientation error over all scenes average.  Indoor, 
it improves the accuracy of $\text{PN}$ by $28\%$ on median position error and $14\%$ on median orientation error on all scenes average. The performance of MS-T$^{vio}$ is similar.
According to Figure~\ref{fig:errdis}, the indoor VIO pose performs slightly better than that of the outdoor, but our framework achieves better results on outdoor datasets. 
This is due to the APR presenting a low accuracy on indoor datasets. This leads to inaccurate poses being identified as accurate during period alignment (see case (3) in Section~\ref{sec:detect}). As such, calculating the reference pose becomes difficult, leading to further inaccuracies during pose optimization.


\subsection{System efficiency}
We assess the performance of the proposed pipeline on an iPhone 14 Pro Max device  on the setup described in Section~\ref{subsec:mobilear}. We measure each parameter over 200 samples.  The average processing time per image is 37 milliseconds while the average time for PoseNet to infer an image is 39.5 milliseconds.
Therefore, our pipeline can perform absolute camera localization on a mobile device in less than 100 milliseconds. 
The ResNet34-based PoseNet requires only 85 MB for weight storage.
We also measure energy consumption as follows. We start with a battery percentage of 100\% and a battery health of 97\%. After 20 minutes of using our application in the Stairs scene, the battery percentage drops to 98\%. With such low computation cost, storage usage, and energy consumption, our framework is realistically usable for AR applications. 

\section{Conclusion}
This paper introduces VIO-APR, a framework that combines an APR with a local VIO tracking system to improve the accuracy and stability of localization for markerless mobile AR. The VIO evaluates and optimizes the APR's accuracy while the APR  corrects VIO drift, resulting in improved positioning. We introduce a new dataset that includes VIO data. Over this dataset, our framework displays up to 36\% higher position accuracy and 29\% higher orientation accuracy, with a significant reduction of low-accuracy frames.
We also demonstrate \sysname's real-life applicability by implementing it into a mobile AR app. The mobile app performs absolute pose estimation in less than 100\,ms with minimal storage and energy consumption.



\bibliographystyle{abbrv-doi}

\bibliography{template}
\end{document}